\documentclass[journal]{IEEEtran}
\usepackage{amssymb}
\usepackage[figuresright]{rotating}
\usepackage[tableposition=top]{caption}
\usepackage{amsmath,subfigure,epsfig,bbding}
\usepackage{epstopdf}
\usepackage{multicol}
\usepackage{url}
\usepackage{booktabs}
\usepackage{multirow}
\usepackage{threeparttable}
\usepackage{xcolor}
\usepackage{array}
\usepackage{arydshln}
\usepackage[mathscr]{eucal}

\DeclareCaptionFont{blue}{\color{blue}}
\usepackage{makecell}
\usepackage{amsmath}
\usepackage[linesnumbered,ruled,vlined]{algorithm2e}
\usepackage{algpseudocode}
\usepackage{amsmath}
\usepackage[normalem]{ulem}
\usepackage[colorlinks=true,
            linkcolor=black,
            anchorcolor=black, 
            citecolor=black, 
            urlcolor=blue
            ]{hyperref}

\def\hlinew#1{%
  \noalign{\ifnum0=`}\fi\hrule \@height #1 \futurelet
   \reserved@a\@xhline}
\newcommand{\PreserveBackslash}[1]{\let\temp=\\#1\let\\=\temp}
\newcolumntype{C}[1]{>{\PreserveBackslash\centering}p{#1}}
\newcolumntype{R}[1]{>{\PreserveBackslash\raggedleft}p{#1}}
\newcolumntype{L}[1]{>{\PreserveBackslash\raggedright}p{#1}}

\begin{document}

\title{Secure Video Quality Assessment Resisting Adversarial Attacks}

\author{Ao-Xiang~Zhang, 
Yuan-Gen~Wang,  
Yu Ran, 
Weixuan Tang, 
Qingxiao Guan, 
and Chunsheng Yang 

\thanks{
This work was partly supported by the National Natural Science Foundation of China under Grant 62272116 and partly by the Graduate Innovation Ability Training Program of Guangzhou University. (\emph{Corresponding author: Yuan-Gen Wang}.)

A.-X. Zhang, Y. Ran, and Q. Guan are with the School of Computer Science and Cyber Engineering, Guangzhou University, Guangzhou 510006, China (e-mail: zax@e.gzhu.edu.cn; ranyu@e.gzhu.edu.cn; gqx@gzhu.edu.cn).

Y.-G. Wang, W. Tang, and C. Yang are with the School of Artificial Intelligence, Guangzhou University, Guangzhou 510006, China (email: wangyg@gzhu.edu.cn; tweix@gzhu.edu.cn; chunsheng.yang@gzhu.edu.cn).

}}

\maketitle

\begin{abstract}
The exponential surge in video traffic has intensified the imperative for Video Quality Assessment (VQA). Leveraging cutting-edge architectures, current VQA models have achieved human-comparable accuracy. However, recent studies have revealed the vulnerability of existing VQA models against adversarial attacks. To establish a reliable and practical assessment system, a secure VQA model capable of resisting such malicious attacks is urgently demanded. Unfortunately, no attempt has been made to explore this issue. This paper first attempts to investigate general adversarial defense principles, aiming to endow existing VQA models with security. Specifically, we first introduce random spatial grid sampling on the video frame for intra-frame defense. Then, we design pixel-wise randomization through a guardian map, globally neutralizing adversarial perturbations. Meanwhile, we extract temporal information from the video sequence as compensation for inter-frame defense. Building upon these principles, we present a novel VQA framework from a security-oriented perspective, termed SecureVQA. Extensive experiments indicate that SecureVQA sets a new benchmark in security while achieving competitive VQA performance compared with state-of-the-art models. Ablation studies delve deeper into analyzing the principles of SecureVQA, demonstrating their generalization and contributions to the security of leading VQA models. The code is available at https://github.com/GZHU-DVL/SecureVQA. 
\end{abstract}

\begin{IEEEkeywords}
No-reference video quality assessment, adversarial defense, black-box attack, model security. 
\end{IEEEkeywords}

\section{Introduction}

\IEEEPARstart{T}{he} tremendous surge of streaming media platforms like YouTube, Facebook, and TikTok has resulted in an exponential increase in video traffic \cite{Trafficreport}. 
This phenomenon has transformed ordinary individuals into both providers and end-users of video services. There is an essential need for Video Quality Assessment (VQA) to automatically evaluate the quality of videos and detect videos that are not up to standard \cite{TOB1,TOB5}. In general, VQA is categorized into three types based on the availability of the pristine video: Full-Reference VQA (FR-VQA), Reduced-Reference VQA (RR-VQA), and No-Reference VQA (NR-VQA). Due to the characteristics of videos on the Internet, NR-VQA has garnered the widest attention \cite{TOB3,TOB4,TLVQM,VIDEVAL,MDTVSFA}. In recent years, driven by the development of cutting-edge architectures, NR-VQA has experienced rapid advancements and achieved human-comparable performance \cite{ZoomVQA,MDVQA,FASTVQA}.

However, recent studies have revealed that similar to classification models, Quality Assessment (QA) models are also remarkably vulnerable to adversarial attacks \cite{AttackIQA, BMVC_AttackVQA, AttackVQA, Ti-Patch, Gushchin2024, Shukla, Sang}. Among them, \cite{AttackIQA} and \cite{BMVC_AttackVQA} first investigated the adversarial attacks on NR-Image QA (NR-IQA) and NR-VQA models under the white-box setting, respectively. \cite{AttackVQA} explored the robustness of the NR-VQA models under the black-box setting for the first time. These vulnerabilities present significant challenges to the practical application of QA models. To establish a reliable and practical assessment system, a secure VQA model that can resist the threat of adversarial attacks is urgently needed.

Unfortunately, although many defense methods have been proposed to protect classification models from adversarial attacks, to the best of our knowledge, almost no attention has been paid to that of VQA models. Two previous works \cite{Defense_1,Defense_2} were the closest to this topic in scope. However, most of their attempts in \cite{Defense_1,Defense_2} directly apply defense strategies from image classification tasks, like blur and median filter in the intra-branch, to achieve the adversarial robustness on IQA models. There are significant differences between the defenses of VQA and image classification. Firstly, typical defense methods for classification models process input data by only removing its adversarial effect with semantic preserving \cite{Defense1,Defense3}. However, the VQA models focus on both semantic information at the macroscopic scale and detailed quality information at the microscopic scale. Thus, we have to deal with them very carefully in such a strategy to reduce the loss of detailed information and avoid quality degradation. Secondly, hundreds of still images are concatenated to form a dynamic video, and the visual quality of the video is simultaneously influenced by details within the frame and dynamic changes between frames \cite{BVQA-2022,TPQI,CoSTA,AdaDqa}. Therefore, the VQA defense strategy must prevent the loss of both intra-frame and inter-frame information. These characteristics make defending VQA much more challenging.

\begin{figure*}[tp]\centering
\subfigure[]
{\includegraphics[width=3.4in]{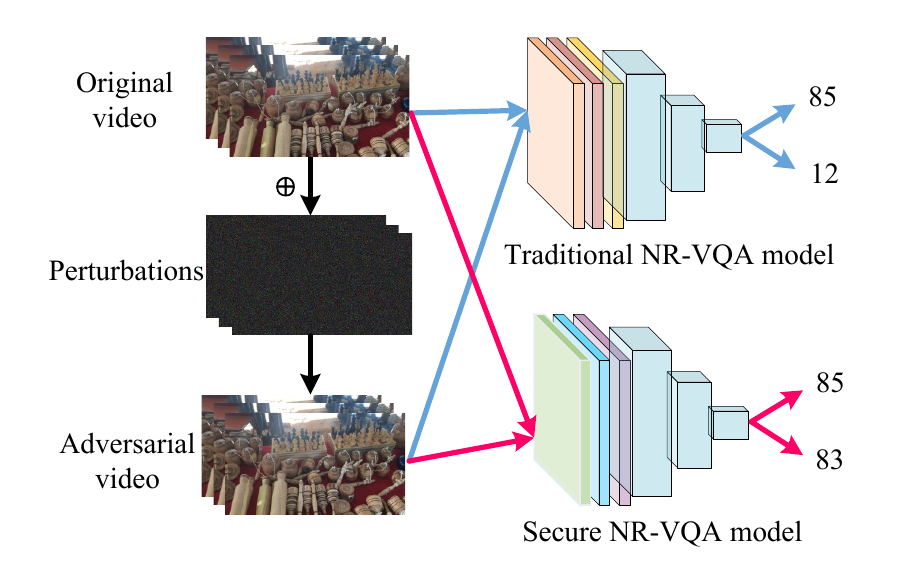}\label{White-score}}
\subfigure[]
{\includegraphics[width=3.3in]{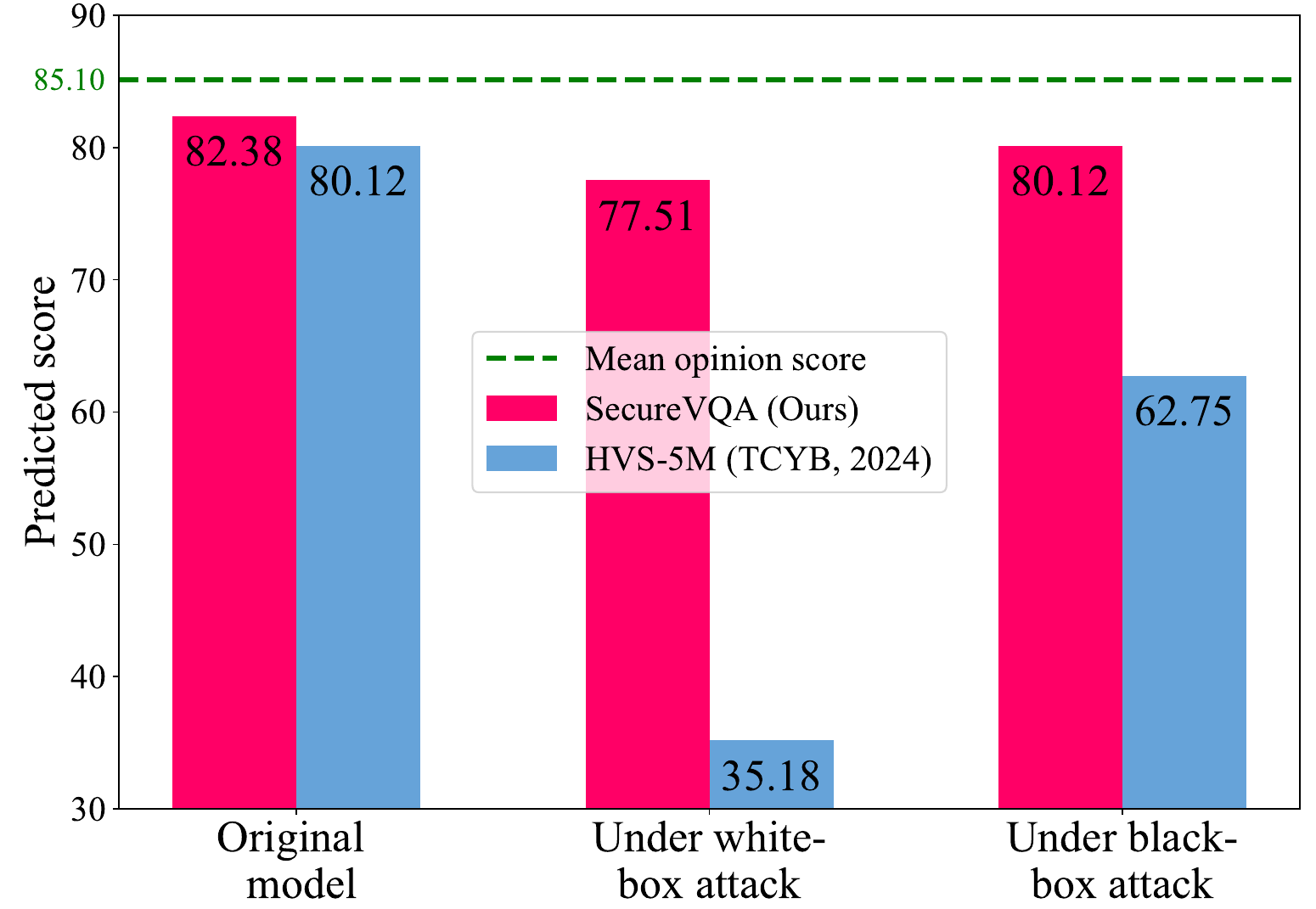}\label{Black-score}}
\caption{The objective of SecureVQA against adversarial attacks. (a) The assessment ability of secure and traditional NR-VQA models on original and adversarial video. (b) The impact of adversarial attacks on SecureVQA (Ours) and HVS-5M \cite{HVS-5M} for the video in (a).}
\label{Definition_SecureVQA}
\end{figure*}

With the above considerations, this paper makes the first attempt to investigate some general adversarial defense principles that can be applied to existing VQA models in a plug-and-play manner and design a robust NR-VQA framework from a security standpoint, termed SecureVQA. 
It is worth noting that SecureVQA does not simply adopt the defense strategies used in classification. Instead, it ingeniously devises solutions tailored to the characteristics of the VQA task. 
In summary, the key idea behind the proposed framework is to weaken the adversarial effect by subtly introducing some randomness and part of additional information as a guarantee. Specifically, SecureVQA designs two branches and applies three defense strategies to each branch for this purpose, each of which can serve as a general component for existing VQA models.

Firstly, we empirically deem randomness as the natural enemy against optimization-based adversarial attacks. Based on it,
we introduce random spatial grid sampling on the video frame for intra-frame defense, which can improve the model's robustness while preserving the detailed information of the video frame. Secondly, a universal strategy termed pixel-wise randomization is proposed. The pixel-wise randomization deliberately introduces random imperceptible modifications through a guardian map to the video before assessment, effectively destroying the carefully crafted perturbations while maximally avoiding the loss of intra-frame information of the video. Overall, the intra-frame defense aims to disrupt adversarial perturbations within individual frames, thereby improving the model's security.
 Thirdly, incorporating additional temporal information is also crucial for model robustness. Given that existing attack methods primarily focus on introducing spatial perturbations, the distinctive temporal information of videos remains unaffected in the intra-frame attacks. Therefore, we introduce inter-frame defense to resist intra-frame attacks. The objective of SecureVQA against adversarial attacks is shown in Fig. \ref{Definition_SecureVQA}. It is anticipated that our framework can not only have strong anti-interference ability for carefully designed adversarial videos but also obtain superior VQA performance. The major contributions of this work are summarized as follows: 
\begin{itemize}
\item We develop three adversarial defense principles, namely spatial grid sampling, pixel-wise randomization, and temporal information extraction, tailored for the characteristics of VQA. These principles are seemingly simple yet quite effective and can be seamlessly integrated into existing VQA models in a plug-and-play manner.
\item We present a general NR-VQA defense framework from the security-oriented viewpoint, termed SecureVQA. Specifically, it splits the adversarial video into two defense branches. The spatial grid sampling and pixel-wise randomization are performed on the video frame for intra-frame defense, while temporal information is extracted as compensation for inter-frame defense. Furthermore, a fusion module is designed to enhance the representation capability of the embeddings from these two branches.
\item Extensive experimental results demonstrate that SecureVQA achieves superior security under both white-box and black-box settings, meanwhile arriving at a competitive VQA performance bar compared with the state-of-the-art. Ablation studies are comprehensively conducted, confirming the contribution of each design to the proposed SecureVQA framework.

\end{itemize}

\begin{figure*}[t]
	\centering
	\includegraphics[width=7.2in]{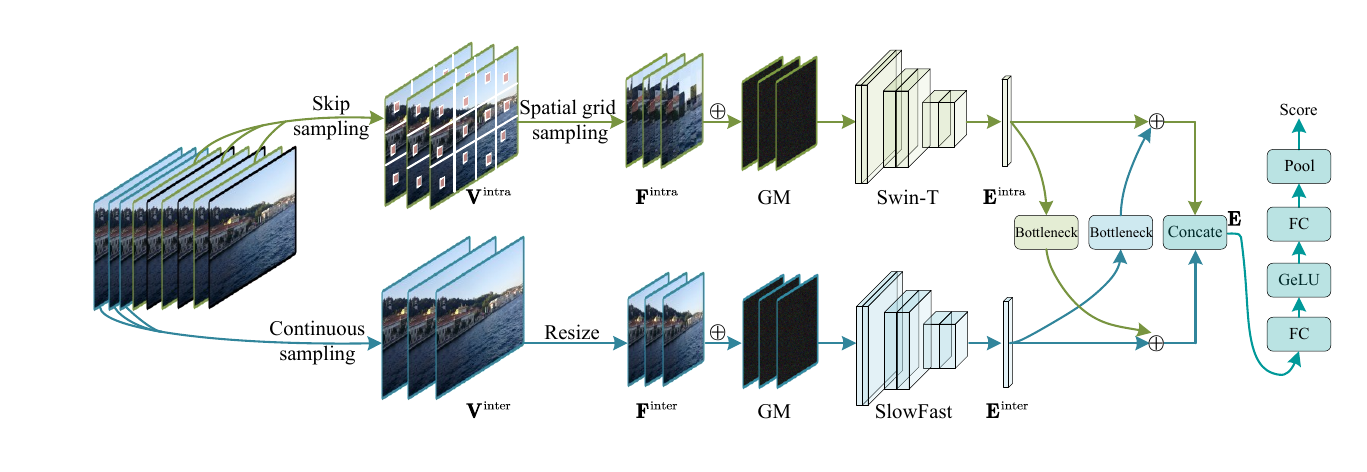}
\vspace{-2.5em}
	\caption{The network architecture of the proposed SecureVQA.
}\label{Framework}
\vspace{-1em}
\end{figure*}

\section{Related work} \label{HVS revisited}
\subsection{NR-VQA Models}
Since \cite{VSFA} first introduced deep learning into NR-VQA, significant progress has been made in improving the capabilities of models from multiple perspectives \cite{Yu2024,Mitra2024,ZoomVQA,MDVQA,FASTVQA}. \cite{TOB2} proposed an NR-VQA method based on spatiotemporal feature fusion and hierarchical information integration through deep neural networks. \cite{SimpleVQA} considered the influence of motion distortion and processed the video into continuous chunks, enabling the extraction of spatial-temporal features from the video. \cite{TiVQA} introduced texture information as a complementary component to content features, aiming to compensate for the human eye's tendency to overlook distortions that occur in regions with intricate textures. \cite{StableVQA} incorporated a blur encoder into the model to detect the blur effect, recognizing the significant influence of video shakiness on quality.  \cite{DiscoVQA} discovered that abrupt scene transitions result in additional degradation in quality, and the Human Visual System (HVS) tends to allocate different levels of attention to frames with varying contents. Subsequently, two Transformer-based modules were designed to handle these two characteristics.  \cite{MaxVQA} systematically divided the video into 13 dimensions to perform a detailed analysis of the individual factors influencing video quality. Additionally, a vision-language-based model was designed to jointly predict specific quality factors and overall quality scores.  \cite{HVS-5M} conducted a thorough survey of HVS, and designed five modules to mimic the five key characteristics of HVS. \cite{DOVER} partitioned the VQA model into technical and aesthetic branches, allowing for a comprehensive evaluation of video quality from two distinct perspectives. \cite{Q-Bench} explored the potential of Large Language Models (LLMs) in VQA tasks for the first time. However, the results indicated that it was extremely difficult for LLMs to output accurate scores directly. Subsequently, \cite{Q-Align} transformed score regression into a classification problem and used the weighted sum as the output scores, validating the ability of LLMs in VQA.

\subsection{Adversarial Attacks on QA models}
In the past two years, some studies have revealed that, similar to classification models, QA models are also vulnerable to adversarial attacks \cite{AttackIQA, BMVC_AttackVQA,AttackVQA, Gushchin2024, Shukla, Sang}. \cite{AttackIQA} made the first attempt to attack the NR-IQA models under the white-box setting. The perturbation was randomly initialized from a discrete set. During the optimization process, the steepest ascent method was utilized to calculate the gradient and update the perturbation at each iteration. To ensure the visual quality of the adversarial images, \cite{AttackIQA} performed human subjective experiments to select the adversarial image that is most similar to the original one. \cite{BMVC_AttackVQA} aimed to increase the quality scores of all QA models by training a fixed-size universal perturbation. In their loss function, a normalization factor was introduced to balance the trade-off between the attack effect and the visual quality. \cite{AttackVQA} conducted the first endeavor to attack the NR-VQA models under the black-box setting, where the attack problem was formulated as misleading the adversarial video's estimated quality score far away from its ground-truth score towards a specific boundary. Recognizing the time-consuming and labor-intensive human subjective experiments, they employed objective metrics $L_2$ and $L_\infty$ norm to constrain the magnitude of perturbation automatically. Besides, \cite{Ranyu} proposed a black-box adversarial attack against NR-IQA models through a random search method in a greedy way. \cite{Yang2024} proposed the correlation-error-based attack framework, which specifically perturbed the score variation on individual images and the correlation within the image set. \cite{AttackIQA_CSVT} proposed a query-based black-box attack method against NR-IQA models with adaptive iterative attacks guided by the initial attack direction.

\section{Proposed framework} \label{proposed method}
In this section, we present the overall pipeline of SecureVQA. Firstly, we divide the video into two branches and apply different types of strategies to each branch for intra and inter-frame defense, respectively. Among them, the intra-frame defense (Section \ref{Spatial defense}) introduces randomness through spatial grid sampling and pixel-wise randomization on the video frame, which aims to disrupt the carefully crafted perturbations and minimize the loss of intra-frame information. The inter-frame defense (Section \ref{Temporal defense}) extracts temporal information as compensation to improve the robustness while avoiding additional modifications to the inter-frame information. After extracting embeddings from the videos in both branches, we fuse the spatial and temporal information (Section \ref{Spatial-Temporal}) to enhance the representation capability of the embeddings. Finally, the quality score of the video can be obtained through simple linear layers.

\subsection{Intra-frame defense} \label{Spatial defense}
In this branch, two principles are designed from the perspective of introducing randomness for intra-frame defense to counter the threat of adversarial attacks. We first perform skip sampling to process the input video ($\mathbf{V}$) into an intra-frame ($\mathbf{V}^{\text{intra}}$) branch, which aims to handle the intra-frame information. The skip sampling process can be formulated as
\begin{equation}\label{nnn}
\mathbf{V}^{\text{intra}}=\mathbf{V}\left[ s : s + n \times d: n \text{, 0 : $H-1$, 0 : $W-1$} \right], 
\end{equation}
where $s : s + n \times d: n$ denotes sampling one frame of every interval $n$ from the starting frame $s$. $d$ denotes the number of frames sampled for processing intra-frame information. $H$ and $W$ denote the height and width of the video frame, respectively.

\textbf{Spatial grid sampling:} To preserve the detailed information and local quality from different regions, spatial sampling is employed to introduce randomness and improve the training efficiency for the intra-frame branch. The main steps are described as follows.
Firstly, the $t$-th frame of $\mathbf{V}^{\text{intra}}$, denoted as $\mathbf{V}^{\text{intra}}_{t}$, is cut into $G \times G$ grids with the same sizes. The grid $\mathbf{g}_{t}^{i,j}$ can be obtained by
\begin{equation}\label{GGG}
    \mathbf{g}_{t}^{i,j}  = \mathbf{V}_{t}^{\text{intra}} \left[ \frac{i\times H}{G}:\frac{\left( i+1 \right) \times H}{G} ,\,\,\frac{j\times W}{G}:\frac{\left( j+1 \right) \times W}{G} \right], 
\end{equation}
where $i$ and $j$ denote the $i$-th row and $j$-th column, respectively. Secondly, a patch of $S \times S$ size is selected from each partitioned grid by 
\begin{equation}\label{SSS}
\mathbf{f}_{t}^{i,j}=\mathbf{g}_{t}^{i,j}\left[ h\ :\ h+S,\ w\ :\ w+S \right],
\end{equation}
where $h$ and $w$ respectively represent the starting height and width of the selected patch in a grid. The regions from different frames are aligned to preserve temporal quality. In this way, we get $G \times G$ patches in a frame. Finally, all the selected patches in a frame are spliced to form the fragments $\mathbf{F}_t^{\text{intra}}$ by
\begin{equation}\label{}
\mathbf{F}_t^{\text{intra}}=\text{Splice}\left( \mathbf{f}_{t}^{0,0}, ..., \mathbf{f}_{t}^{G-1,G-1}  \right). 
\end{equation}

\textbf{Pixel-wise randomization:} We design a general defense strategy named pixel-wise randomization, which aims to neutralize the adversarial effect and provide global immunity to the model by introducing random and subtle modifications on the video frames. Specifically, a guardian map ($\mathbf{\text{GM}}$) with the same dimensions as $\mathbf{F}^{\text{intra}}$ is randomly and uniformly initialized, whose element is independently sampled from a discrete set $\left\{-1, +1\right\}$. Afterwards, $\mathbf{\text{GM}}$ is added to $\mathbf{F}^{\text{intra}}$ in an element-wise manner by 
\begin{equation}\label{}
\mathbf{F}_{t}^{\text{intra}}=\mathbf{F}_{t}^{\text{intra}} + \mathbf{\text{GM}} ,
\end{equation}
which can effectively eliminate the carefully crafted adversarial perturbation while preserving the visual quality of the defense video with minimal alterations. After applying the spatial sampling and pixel-wise randomization strategies, the input video $\mathbf{V}$ is transformed into $\mathbf{F}^{\text{intra}}$ in the intra-frame branch, and then $\mathbf{F}^{\text{intra}}$ is fed into video Swin-Transformer (Swin-T) to obtain intra-embeddings $\mathbf{E}^{\text{intra}}$ as 
\begin{equation}\label{}
\mathbf{E}^{\text{intra}}=\text{Swin-T}\left( \mathbf{F}^{\text{intra}} \right) ,
\end{equation} 
which leverages attention mechanisms to integrate spatial information across the patches selected in each partitioned grid. This allows for the interaction of local quality among different regions, thereby determining the overall quality of the video.

\subsection{Inter-frame defense} \label{Temporal defense}
In this branch, we extract temporal information as compensation for inter-frame defense to counter the threat of adversarial attacks. Firstly, we perform the continuous sampling to process the input video ($\mathbf{V}$) into an inter-frame ($\mathbf{V}^{\text{inter}}$) branch, and the continuous sampling process can be formulated as
\begin{equation}\label{ddd}
\mathbf{V}^{\text{inter}}=\mathbf{V}\left[ s : s + d \text{, 0 : $H-1$, 0 : $W-1$} \right], 
\end{equation}
where $s : s + d$ denotes selecting $d$ consecutive frames for extracting temporal information, and $s$ is set to a different value from the intra-frame branch. Considering that the details and high-frequency noise of the video frame are preserved through spatial grid sampling in the intra-frame branch, we therefore employ the resize operation as a supplementary in the inter-frame branch to comprehensively extract video embeddings by
\begin{equation}\label{resize}
\mathbf{F}_t^{\text{inter}}=\text{Resize}\left( \mathbf{V}_t^{\text{inter}}  \right). 
\end{equation}

Similarly, the pixel-wise randomization is also adopted through the guardian map to introduce randomness and neutralize the meticulously crafted perturbation due to its minimal impact on performance and significant contribution to security by
\begin{equation}\label{}
\mathbf{F}_{t}^{\text{inter}}=\mathbf{F}_{t}^{\text{inter}} + \mathbf{\text{GM}} .
\end{equation}

\textbf{Temporal information extraction:} After applying the resize operation and pixel-wise randomization strategies, the input video $\mathbf{V}$ is transformed into $\mathbf{F}^{\text{inter}}$ in the inter-frame branch, and then $\mathbf{F}^{\text{inter}}$ is fed into SlowFast by
\begin{equation}\label{}
\mathbf{E}^{\text{inter}}=\text{SlowFast}\left( \mathbf{F}^{\text{inter}} \right) ,
\end{equation}
which effectively captures rapidly changing temporal information without computing the optical flow. SlowFast operates with slow and fast pathways, which are responsible for extracting semantic and temporal information, respectively. Here, we only adopt the embeddings extracted by the fast path as inter-frame embeddings $\mathbf{E}^{\text{inter}}$. 



\begin{table*}[]
\centering
\caption{Performance evaluations under white-box setting on VSFA \cite{VSFA}, MDTVSFA \cite{MDTVSFA}, TiVQA \cite{TiVQA}, BVQA-2022 \cite{BVQA-2022}, and HVS-5M \cite{HVS-5M}. Here, the performance before the attack is marked in gray. }\label{white-box}
\renewcommand\arraystretch{1.33}
\fontsize{9}{9}\selectfont
\setlength{\tabcolsep}{0.10mm}{
\begin{tabular}{c:cccc:cccc:cccc:cccc:cccc}
\hline
Dataset                    &           & \multicolumn{3}{c:}{KoNViD-1k}     &  & \multicolumn{3}{c:}{LIVE-VQC} &  & \multicolumn{3}{c:}{YouTube-UGC} &  & \multicolumn{3}{c:}{$\text{LSVQ}_\text{test}$} &  & \multicolumn{3}{c}{$\text{LSVQ}_\text{1080P}$} \\  \cmidrule(lr){3-5} \cmidrule(lr){7-9} \cmidrule(lr){11-13} \cmidrule(lr){15-17} \cmidrule(lr){19-21} 
Metric                     &           & SRCC      & PLCC      & $R$         &  & SRCC      & PLCC     & $R$     &  & SRCC       & PLCC      & $R$     &  & SRCC      & PLCC      & $R$      &  & SRCC       & PLCC      & $R$      \\ \hline
\multirow{2}{*}{VSFA}      &           & -0.7155 & -0.8139 & -0.2521    &  & -0.7526 & -0.7323 & -0.4968    &  & -0.7013  & -0.8025 & -0.3669    &  & -0.6601 & -0.6824 & 0.2551     &  & -0.7864  & -0.7833 & 0.0528     \\
                           &  & \textcolor{gray}{0.7882}  & \textcolor{gray}{0.8106}  & \textcolor{gray}{-} &  & \textcolor{gray}{0.7665}  & \textcolor{gray}{0.7506}  & \textcolor{gray}{-} &  & \textcolor{gray}{0.7492}   & \textcolor{gray}{0.7721}  & \textcolor{gray}{-} &  & \textcolor{gray}{0.7865}  & \textcolor{gray}{0.7972}  & \textcolor{gray}{-} &  & \textcolor{gray}{0.7040}   & \textcolor{gray}{0.6868}  & \textcolor{gray}{-} \\ \hline
\multirow{2}{*}{MDTVSFA}   &           & -0.7296 & -0.8219 & -0.0419    &  & -0.7766 & -0.7720 & -0.3129    &  & -0.7238  & -0.7946 & -0.2974    &  & -0.6569 & -0.7949 & 0.4086     &  & -0.7438  & -0.7878 & 0.2533     \\
                           &           & \textcolor{gray}{0.8003}  & \textcolor{gray}{0.8074}  & \textcolor{gray}{-}          &  & \textcolor{gray}{0.7908}  & \textcolor{gray}{0.8091}  & \textcolor{gray}{-}          &  & \textcolor{gray}{0.7683}   & \textcolor{gray}{0.7950}  & \textcolor{gray}{-}          &  & \textcolor{gray}{0.8136}  & \textcolor{gray}{0.8125}  & \textcolor{gray}{-}          &  & \textcolor{gray}{0.7114}   & \textcolor{gray}{0.7009}  & \textcolor{gray}{-}          \\ \hline
\multirow{2}{*}{TiVQA}     &           & -0.7419 & -0.8230 & -0.2521    &  & -0.6833 & -0.7425 & -0.2214    &  & -0.6034  & -0.7199 & 0.0479     &  & -0.7591 & -0.7935 & 0.0178     &  & -0.7351  & -0.7442 & 0.1875     \\
                           &           & \textcolor{gray}{0.8046}  & \textcolor{gray}{0.8377}  & \textcolor{gray}{-}          &  & \textcolor{gray}{0.8179}  & \textcolor{gray}{0.8067}  & \textcolor{gray}{-}          &  & \textcolor{gray}{0.7907}   & \textcolor{gray}{0.8177}  & \textcolor{gray}{-}          &  & \textcolor{gray}{0.8331}  & \textcolor{gray}{0.8343}  & \textcolor{gray}{-}          &  & \textcolor{gray}{0.7478}   & \textcolor{gray}{0.7670}  & \textcolor{gray}{-}          \\ \hline
\multirow{2}{*}{BVQA-2022} &           & -0.7265 & -0.8224 & -0.0708    &  & -0.6712 & -0.6867 & -0.0229    &  & -0.6920  & -0.6889 & 0.0817     &  & -0.7769 & -0.8093 & 0.0533     &  & -0.7128  & -0.7613 & 0.1876     \\
                           &           & \textcolor{gray}{0.8427}  & \textcolor{gray}{0.8469}  & \textcolor{gray}{-}          &  & \textcolor{gray}{0.8698}  & \textcolor{gray}{0.8457}  & \textcolor{gray}{-}          &  & \textcolor{gray}{0.8145}   & \textcolor{gray}{0.8500}  & \textcolor{gray}{-}          &  & \textcolor{gray}{0.8613}  & \textcolor{gray}{0.8531}  & \textcolor{gray}{-}          &  & \textcolor{gray}{0.7875}   & \textcolor{gray}{0.7799}  & \textcolor{gray}{-}          \\ \hline
\multirow{2}{*}{HVS-5M} &           & -0.7833 & -0.8218 & 0.1034    &  & -0.7149 & -0.7666 & -0.1614    &  & -0.6537  & -0.7858 & -0.1063     &  & -0.7041 & -0.7918 & 0.5719     &  & -0.7377  & -0.7685 & 0.3246     \\
                           &           & \textcolor{gray}{0.8715}  & \textcolor{gray}{0.8605}  & \textcolor{gray}{-}          &  & \textcolor{gray}{0.8662}  & \textcolor{gray}{0.8802}  & \textcolor{gray}{-}          &  & \textbf{\textcolor{gray}{0.8640}}   & \textbf{\textcolor{gray}{0.8612}}  & \textcolor{gray}{-}          &  & \textcolor{gray}{0.8866}  & \textcolor{gray}{0.8809}  & \textcolor{gray}{-}          &  & \textcolor{gray}{0.7901}   & \textcolor{gray}{0.7947}  & \textcolor{gray}{-}          \\ \hline
\multirow{2}{*}{SecureVQA} &           &  \textbf{-0.1865}         &  \textbf{-0.1311}         &          \textbf{1.6743} &  &     \textbf{0.5923}      &     \textbf{0.5103}     &    \textbf{2.0843}   &  &    \textbf{0.2034}        &           \textbf{0.1531} &    \textbf{0.9954}    &  &    \textbf{-0.0861}        &           \textbf{-0.0264} &    \textbf{1.6225}    &  &            \textbf{0.5774}       &   \textbf{0.6660}        &    \textbf{3.0144} \\
                           &           & \textbf{\textcolor{gray}{0.8805}}          &          \textbf{\textcolor{gray}{0.8692}} &    \textcolor{gray}{-}                &  &           \textbf{\textcolor{gray}{0.8861}}          &          \textbf{\textcolor{gray}{0.8959}} &    \textcolor{gray}{-}       &  &            \textcolor{gray}{0.8542}          &          \textcolor{gray}{0.8369} &    \textcolor{gray}{-}        &  &           \textbf{\textcolor{gray}{0.9008}}          &          \textbf{\textcolor{gray}{0.9087}} &    \textcolor{gray}{-}        &  &            \textbf{\textcolor{gray}{0.7949}}          &          \textbf{\textcolor{gray}{0.8305}} &    \textcolor{gray}{-}        \\ \hline
\end{tabular}}
\end{table*}
\begin{table*}[]
\centering
\caption{Performance evaluations under black-box setting on VSFA \cite{VSFA}, MDTVSFA \cite{MDTVSFA}, TiVQA \cite{TiVQA}, BVQA-2022 \cite{BVQA-2022}, and HVS-5M \cite{HVS-5M}. Here, the performance before the attack is marked in gray. }\label{black-box}
\renewcommand\arraystretch{1.33}
\fontsize{9}{9}\selectfont
\setlength{\tabcolsep}{0.1mm}{
\begin{tabular}{c:cccc:cccc:cccc:cccc:cccc}
\hline
Dataset                    &           & \multicolumn{3}{c:}{KoNViD-1k}     &  & \multicolumn{3}{c:}{LIVE-VQC} &  & \multicolumn{3}{c:}{YouTube-UGC} &  & \multicolumn{3}{c:}{$\text{LSVQ}_\text{test}$} &  & \multicolumn{3}{c}{$\text{LSVQ}_\text{1080P}$} \\  \cmidrule(lr){3-5} \cmidrule(lr){7-9} \cmidrule(lr){11-13} \cmidrule(lr){15-17} \cmidrule(lr){19-21} 
Metric                     &           & SRCC      & PLCC      & $R$         &  & SRCC      & PLCC     & $R$     &  & SRCC       & PLCC      & $R$     &  & SRCC      & PLCC      & $R$      &  & SRCC       & PLCC      & $R$      \\ \hline
\multirow{2}{*}{VSFA}      &           & -0.0305 & 0.0586  & 1.6573     &  & -0.1605 & -0.0132 & 1.3233     &  & -0.2231  & -0.1017  & 1.5499     &  & 0.2386   & 0.2413  & 2.3545     &  & 0.2351   & 0.2840  & 2.1839     \\
                           &  & \textcolor{gray}{0.7882}  & \textcolor{gray}{0.8106}  & \textcolor{gray}{-} &  & \textcolor{gray}{0.7665}  & \textcolor{gray}{0.7506}  & \textcolor{gray}{-} &  & \textcolor{gray}{0.7492}   & \textcolor{gray}{0.7721}  & \textcolor{gray}{-} &  & \textcolor{gray}{0.7865}  & \textcolor{gray}{0.7972}  & \textcolor{gray}{-} &  & \textcolor{gray}{0.7040}   & \textcolor{gray}{0.6868}  & \textcolor{gray}{-} \\ \hline
\multirow{2}{*}{MDTVSFA}   &           & 0.0261  & 0.1235  & 1.7587     &  & -0.0074 & 0.0807  & 1.4391     &  & -0.0646  & 0.0532   & 1.5080     &  & -0.0706  & -0.0384 & 2.1352     &  & -0.1339  & -0.0288 & 1.8592     \\
                           &           & \textcolor{gray}{0.8003}  & \textcolor{gray}{0.8074}  & \textcolor{gray}{-}          &  & \textcolor{gray}{0.7908}  & \textcolor{gray}{0.8091}  & \textcolor{gray}{-}          &  & \textcolor{gray}{0.7683}   & \textcolor{gray}{0.7950}  & \textcolor{gray}{-}          &  & \textcolor{gray}{0.8136}  & \textcolor{gray}{0.8125}  & \textcolor{gray}{-}          &  & \textcolor{gray}{0.7114}   & \textcolor{gray}{0.7009}  & \textcolor{gray}{-}          \\ \hline
\multirow{2}{*}{TiVQA}     &           & -0.4992 & -0.4099 & 1.2458     &  & -0.0762 & 0.0452  & 1.3234     &  & -0.1357  & -0.0087  & 1.3478     &  & -0.3343  & -0.2161 & 1.8086     &  & -0.1414  & -0.0343 & 1.3327     \\
                           &           & \textcolor{gray}{0.8046}  & \textcolor{gray}{0.8377}  & \textcolor{gray}{-}          &  & \textcolor{gray}{0.8179}  & \textcolor{gray}{0.8067}  & \textcolor{gray}{-}          &  & \textcolor{gray}{0.7907}   & \textcolor{gray}{0.8177}  & \textcolor{gray}{-}          &  & \textcolor{gray}{0.8331}  & \textcolor{gray}{0.8343}  & \textcolor{gray}{-}          &  & \textcolor{gray}{0.7478}   & \textcolor{gray}{0.7670}  & \textcolor{gray}{-}          \\ \hline
\multirow{2}{*}{BVQA-2022} &           & 0.1978  & 0.2215  & 2.0480     &  & 0.4571  & 0.4778  & 2.2061     &  & 0.3951   & 0.4306   & 2.1072     &  & 0.3355   & 0.4240  & 2.2269     &  & 0.3917   & 0.4678  & 2.0796     \\
                           &           & \textcolor{gray}{0.8427}  & \textcolor{gray}{0.8469}  & \textcolor{gray}{-}          &  & \textcolor{gray}{0.8698}  & \textcolor{gray}{0.8457}  & \textcolor{gray}{-}          &  & \textcolor{gray}{0.8145}   & \textcolor{gray}{0.8500}  & \textcolor{gray}{-}          &  & \textcolor{gray}{0.8613}  & \textcolor{gray}{0.8531}  & \textcolor{gray}{-}          &  & \textcolor{gray}{0.7875}   & \textcolor{gray}{0.7799}  & \textcolor{gray}{-}          \\ \hline
\multirow{2}{*}{HVS-5M} &           & -0.1207 & -0.0404 & 1.5116    &  & 0.1948 & 0.3045 & 1.6168    &  & -0.0326 & 0.1086 & 1.5243     &  & 0.1552 & 0.3075 & 1.8900     &  & 0.2290  & 0.3181 & 2.2802     \\
                           &           & \textcolor{gray}{0.8715}  & \textcolor{gray}{0.8605}  & \textcolor{gray}{-}          &  & \textcolor{gray}{0.8662}  & \textcolor{gray}{0.8802}  & \textcolor{gray}{-}          &  & \textbf{\textcolor{gray}{0.8640}}   & \textbf{\textcolor{gray}{0.8612}}  & \textcolor{gray}{-}          &  & \textcolor{gray}{0.8866}  & \textcolor{gray}{0.8809}  & \textcolor{gray}{-}          &  & \textcolor{gray}{0.7901}   & \textcolor{gray}{0.7947}  & \textcolor{gray}{-}          \\ \hline
\multirow{2}{*}{SecureVQA} &           &       \textbf{0.8788}     &     \textbf{0.8590}       & \textbf{5.7787}           &  &       \textbf{0.8635}     &  \textbf{0.8669}         &  \textbf{4.2794}     &  &       \textbf{0.8512}     &     \textbf{0.8432}      &   \textbf{4.6186}     &  &      \textbf{0.8969}     &    \textbf{0.9074}       &   \textbf{5.5061}     &  &           \textbf{0.7918} &       \textbf{0.8288}    &   \textbf{4.2826}     \\
                           &           &           \textbf{\textcolor{gray}{0.8805}}          &          \textbf{\textcolor{gray}{0.8692}} &    \textcolor{gray}{-}           &  &           \textbf{\textcolor{gray}{0.8861}}          &          \textbf{\textcolor{gray}{0.8959}} &    \textcolor{gray}{-}       &  &            \textcolor{gray}{0.8542}          &          \textcolor{gray}{0.8369} &    \textcolor{gray}{-}        &  &           \textbf{\textcolor{gray}{0.9008}}          &          \textbf{\textcolor{gray}{0.9087}} &    \textcolor{gray}{-}        &  &            \textbf{\textcolor{gray}{0.7949}}          &          \textbf{\textcolor{gray}{0.8305}} &    \textcolor{gray}{-}        \\ \hline
\end{tabular}}
\end{table*}
\noindent

\subsection{Spatial-temporal information fusion} \label{Spatial-Temporal}
Considering that the embeddings of the two branches are respectively extracted by Transformer and Convolutional Neural Networks, we thus design an information fusion module to integrate the embeddings from these two distinct domains instead of simply concatenation, aiming at enhancing the overall representation capability of the embeddings. The fusion process can be expressed as
\begin{equation}\label{}
\mathbf{E}_{\text{fuse}}^{\text{intra}}=\mathbf{E}^{\text{intra}}+\text{Bottleneck}\left( \mathbf{E}^{\text{inter}} \right), 
\end{equation}
\begin{equation}\label{}
\mathbf{E}_{\text{fuse}}^{\text{inter}}=\mathbf{E}^{\text{inter}}+\text{Bottleneck}\left( \mathbf{E}^{\text{intra}} \right), 
\end{equation}
where the $\text{Bottleneck}$ operation can be expressed as three fully-connected (FC) networks as
\begin{equation}\label{}
\widehat{\mathbf{E}}^{\text{intra/inter}}=\text{FC}\left( \text{ReLU}\left( \text{FC}\left( \text{ReLU}\left( \text{FC}\left( \mathbf{E}^{\text{intra/inter}} \right) \right) \right) \right) \right). 
\end{equation}

Next, the embeddings from the above two branches after fusion are concatenated (Concate) to form the robust embeddings $\mathbf{E}$ of the input video $\mathbf{V}$ as
\begin{equation}\label{}
\mathbf{E} =\text{Concate}\left( \mathbf{E}_{\text{fuse}}^{\text{intra}}, \mathbf{E}_{\text{fuse}}^{\text{inter}} \right) .
\end{equation}

Finally, the robust embeddings $\mathbf{E}$ are processed by $\text{FC}$ and average pooling  ($\text{Pool}$) to obtain the prediction score of the video as
\begin{equation}\label{}
\text{score}=\text{Pool}\left( \text{FC}\left( \text{GeLU}\left( \text{FC}\left( \mathbf{E} \right) \right) \right) \right) .
\end{equation}

\section{Experimental results and analysis}
\subsection{Experimental setup}

\textbf{Test datasets:}
We conduct experiments to evaluate the security and accuracy of the proposed SecureVQA on four mainstream VQA datasets, including KoNViD-1k~\cite{KoNViD}, LIVE-VQC~\cite{VQC}, YouTube-UGC~\cite{YouTube}, and LSVQ~\cite{LSVQ}. The large-scale LSVQ is further divided into a training set ($\text{LSVQ}_\text{train}$) and two testing sets ($\text{LSVQ}_\text{test}$ and $\text{LSVQ}_\text{1080P}$). Among them, $\text{LSVQ}_\text{test}$ contains videos from 240P to 720P, while $\text{LSVQ}_\text{1080P}$ includes high-resolution 1080P videos.

\textbf{Evaluation metrics and loss function:}
The security and VQA performance of models are evaluated by Spearman Rank order Correlation Coefficient (SRCC) and Pearson’s Linear Correlation Coefficient (PLCC). SRCC and PLCC evaluate the monotonic and linear relationship between the predicted score and the Mean Opinion Score (MOS), respectively. To evaluate the security of models, we follow~\cite{AttackIQA} to compute the actual deviation of a single video after being attacked, which is formulated as 
\begin{equation} \label{R_values}
R=\frac{1}{K}\sum_{i=1}^K{\log \left( \frac{\,\,\left| f\big( \langle \mathbf{V} \rangle _i \big) -\text{Tar}\big(  \langle \mathbf{V} \rangle _i \big) \right|}{\left| f\big( \langle \mathbf{V} \rangle _i \big) -f\big( \langle \mathbf{V} \rangle _{i}^{\text{adv}} \big) \right|} \right) ,}
\end{equation}
where $K$ denotes the number of videos in the testing set, $\langle \mathbf{V} \rangle _i$ and $\langle \mathbf{V} \rangle _{i}^{\text{adv}}$ denote the $i$-th original and adversarial videos, respectively. $f \left( \cdot \right)$ denotes the NR-VQA model, and $\text{Tar} \left( \cdot \right)$ is a predetermined value deviating from its MOS, indicating the target score expected for the original video $\langle \mathbf{V} \rangle _i$ after being attacked. A higher value of $R$ indicates better model robustness. The loss function is formulated as the sum of PLCC and approximately differentiable SRCC \cite{SRCC}.

\begin{figure*}[t]
	\centering
	\includegraphics[width=6.2in]{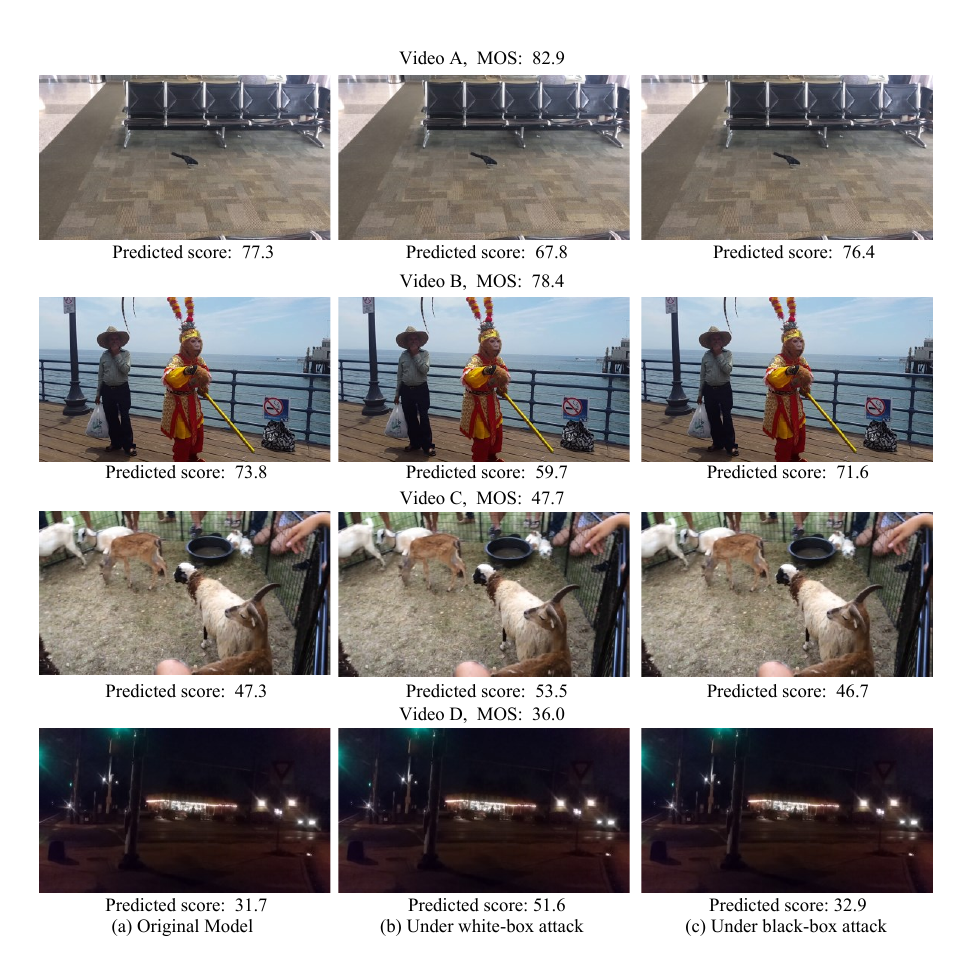}
	\caption{SecureVQA predicted scores for original videos and corresponding adversarial videos.
}\label{Effect_picture}
\vspace{-1em}
\end{figure*}

\textbf{Attack strategy:} Currently, research on the security of VQA is in its preliminary stage. Most of the related work focus on adversarial attacks on IQA. As far as we know, \cite{AttackVQA} is the only representative adversarial attack on VQA under the black box setting. Therefore, we select \cite{AttackVQA} as the white-box and black-box attack benchmark in this paper, which designs an efficient way to address the computation issue within adversarial video. Specifically, \cite{AttackVQA} employs the Projected Gradient Descent (PGD) \cite{PGD} algorithm for white-box attack, restricting the pixel-wise $L_2$ or $L_{\infty}$ norm of the perturbations to 1/255 or 3/255 in each iteration. In a black-box attack, \cite{AttackVQA} tentatively perturbs a portion of the frame (56$\times$56 pixels) while leaving the remainder untouched at each query. Such perturbations would be kept if the model's score changes in the expected directions, and abandoned otherwise. After 300 times of queries, the resulting adversarial frame is successfully obtained. Besides, considering the computational complexity, \cite{AttackVQA} selects 50 videos in the dataset to verify the model security. In this paper, we adopt the same setting as \cite{AttackVQA} for fair comparison.     
\noindent

\textbf{Implementation details:}
The experiments are built on Pytorch 3.8 and run on one RTX 4090 GPU card. In the two branches that handle the intra-frame and inter-frame information, the Swin-Transformer~\cite{Swin} pre-trained on ImageNet~\cite{ImageNet} and SlowFast~\cite{Slowfast} pre-trained on Kinetics~\cite{Kinetics} are utilized as backbone networks to extract embeddings. Wherein, the parameters of Swin-Transformer are set to be trainable, and the parameters of SlowFast are set to be frozen due to its large amount of computation. Such a training setup can speed up the training process. In the skip sampling and continuous sampling process, $n$ and $d$ (in Equations \ref{nnn} and \ref{ddd}) are set to 2 and 32 respectively, and $\mathbf{V}_t^{\text{inter}}$ is resized to 224 $\times$ 224 (in Equation \ref{resize}). In addition, the inter-frame branch splits the video into 16 segments. In the spatial sampling process, $G$ and $S$ (in Equations \ref{GGG} and \ref{SSS}) are set to 7 and 32, respectively, as done in \cite{FASTVQA}. The Adam optimizer with the initial learning rate of 0.001 and batch size 12 is used for training.


\begin{table*}[t] 
\centering
\fontsize{9}{9}\selectfont
\setlength{\tabcolsep}{2.75mm}{
\renewcommand\arraystretch{1.3}
\caption{Evaluations on the large-scale dataset. Here, the top three performers are highlighted in boldface, underlined, and wavy line, respectively, in descending order of performance.}\label{LSVQ}
\begin{tabular}{l:cccccccccccc}
\hline
Training Set              & \multicolumn{8}{c}{$\text{LSVQ}_\text{train}$}                                                                                                              \\ \hline
Testing Set               & \multicolumn{2}{c:}{$\text{LSVQ}_\text{test}$}     & \multicolumn{2}{c:}{$\text{LSVQ}_\text{1080P}$}     & \multicolumn{2}{c:}{KoNViD-1k}     & \multicolumn{2}{c}{LIVE-VQC} \\ \hline
Metric                    & SRCC  & \multicolumn{1}{c:}{PLCC}  & SRCC  & \multicolumn{1}{c:}{PLCC}  & SRCC  & \multicolumn{1}{c:}{PLCC}  & SRCC          & PLCC         \\ \hline
TLVQM \cite{TLVQM} (TIP, 2019)         & 0.772 & \multicolumn{1}{c:}{0.774} & 0.589 & \multicolumn{1}{c:}{0.616} & 0.732 & \multicolumn{1}{c:}{0.724} & 0.670         & 0.691        \\
VIDEVAL \cite{VIDEVAL} (TIP, 2021)       & 0.795 & \multicolumn{1}{c:}{0.783} & 0.545 & \multicolumn{1}{c:}{0.554} & 0.751 & \multicolumn{1}{c:}{0.741} & 0.630         & 0.640        \\
VSFA \cite{VSFA} (ACM MM, 2019)       & 0.801 & \multicolumn{1}{c:}{0.796} & 0.675 & \multicolumn{1}{c:}{0.704} & 0.784 & \multicolumn{1}{c:}{0.795} & 0.734         & 0.772        \\
$\text{Patch-VQ}_{w/ patch}$ \cite{LSVQ} (CVPR, 2021)  & 0.827 & \multicolumn{1}{c:}{0.828} & 0.711 & \multicolumn{1}{c:}{0.739} & 0.791 & \multicolumn{1}{c:}{0.795} & 0.770         & 0.807        \\
BVQA-2022 \cite{BVQA-2022} (TCSVT, 2022)   & 0.852 & \multicolumn{1}{c:}{0.855} & 0.771 & \multicolumn{1}{c:}{0.782} & 0.834 & \multicolumn{1}{c:}{0.837} & 0.816         & 0.824        \\
FAST-VQA \cite{FASTVQA} (ECCV, 2022)     & 0.876 & \multicolumn{1}{c:}{\uwave{0.877}} & 0.779 & \multicolumn{1}{c:}{0.814} & \uwave{0.859} & \multicolumn{1}{c:}{\uwave{0.855}} & \uwave{0.823}         & \uwave{0.844}        \\
DiscoVQA \cite{DiscoVQA} (TCSVT, 2023)     & 0.859 & \multicolumn{1}{c:}{0.850} & N/A & \multicolumn{1}{c:}{N/A} & 0.846 & \multicolumn{1}{c:}{0.849} & 0.823         & 0.837        \\
HVS-5M \cite{HVS-5M} (TCYB, 2024)     & \uwave{0.879} & \multicolumn{1}{c:}{0.872} & \textbf{0.798} & \multicolumn{1}{c:}{\uwave{0.815}} & 0.857 & \multicolumn{1}{c:}{0.855} & 0.810         & 0.832        \\
DOVER \cite{DOVER} (ICCV, 2023)        & \textbf{0.888} & \multicolumn{1}{c:}{\textbf{0.889}} & \underline{0.795} & \multicolumn{1}{c:}{\textbf{0.830}} & \textbf{0.884} & \multicolumn{1}{c:}{\textbf{0.883}} & \underline{0.832}         & \underline{0.855}        \\
SecureVQA (Ours)          &  \underline{0.883}     & \multicolumn{1}{c:}{\underline{0.884}}      & \uwave{0.785}      & \multicolumn{1}{c:}{\underline{0.821}}      & \underline{0.868}     & \multicolumn{1}{c:}{\underline{0.865}}      &              \textbf{0.848} &     \textbf{0.859}         \\ \hline
\end{tabular}}
\end{table*}
\begin{table*}[t] 
\centering
\caption{Evaluations on smaller datasets. Here, the top three performers are highlighted in boldface, underlined, and wavy line, respectively, in descending order of performance. }\label{fine-tuning}
\renewcommand\arraystretch{1.3}
\fontsize{9}{9}\selectfont
\setlength{\tabcolsep}{2.35mm}{
\begin{tabular}{l:cc:cc:cc:cc}
\hline
Fine-tuning Dataset     & \multicolumn{2}{c:}{LIVE-VQC (585)} & \multicolumn{2}{c:}{KoNViD-1k (1200)} & \multicolumn{2}{c:}{YouTube-UGC (1380)} & \multicolumn{2}{c}{Weighted Average} \\
Metric                  & SRCC           & PLCC          & SRCC          &  PLCC          & \ \ \ SRCC            & \ \ \ PLCC           & SRCC              & PLCC             \\ \hline
TLVQM \cite{TLVQM} (TIP, 2019)       & 0.799          & 0.803         & 0.773         & 0.768         & \ \ 0.669           & \ \ 0.659          & 0.732             & 0.726            \\
VIDEVAL \cite{VIDEVAL} (TIP, 2021)     & 0.752          & 0.751         & 0.783         & 0.780 & \ \ 0.779           & \ \ 0.773          & 0.772             & 0.772            \\
VSFA \cite{VSFA} (ACM MM, 2019)     & 0.773          & 0.795         & 0.773         & 0.775         & \ \ 0.724           & \ \ 0.743          & 0.752             & 0.765            \\
Patch-VQ \cite{LSVQ} (CVPR, 2021)   & 0.827          & 0.837         & 0.791         & 0.786         & \ \ N/A             & \ \ N/A            & N/A               & N/A              \\
BVQA-2022 \cite{BVQA-2022} (TCSVT, 2022) & 0.834          & 0.842         & 0.834         & 0.836         & \ \ 0.818           & \ \ 0.826          & 0.823             & 0.833            \\
FAST-VQA \cite{FASTVQA} (ECCV, 2022)   & 0.849          & 0.862         & \underline{0.891}         & \underline{0.892}         & \ \ 0.855           & \ \ 0.852          & 0.868             & 0.869            \\
DiscoVQA \cite{DiscoVQA} (TCSVT, 2023)  & 0.820           & 0.826         & 0.846         & 0.849         & \ \ 0.809           & \ \ 0.808          & 0.820              & 0.822            \\
HVS-5M \cite{HVS-5M} (TCYB, 2024)     &        \underline{0.878}        &    \underline{0.879}           &         0.882      &       0.882        &          \ \  \underline{0.880} &  \ \   \underline{0.878}           &      \underline{0.880}             &     \underline{0.880}             \\
DOVER \cite{DOVER} (ICCV, 2023)      & \uwave{0.860}           & \uwave{0.875}         & \textbf{0.909}         & \textbf{0.906}         & \ \ \textbf{0.890}            & \ \ \textbf{0.891}          & \textbf{0.891}             & \textbf{0.891}            \\
SecureVQA (Ours)        & \textbf{0.882}                &  \textbf{0.880}             &     \uwave{0.885}          &     \uwave{0.883}          &    \ \   \uwave{0.862}          &     \ \   \uwave{0.860}                 &       \uwave{0.874}            &                 \uwave{0.872} \\ \hline
\end{tabular}}
\end{table*}

\subsection{Evaluations on model security} \label{individual datasets}
In this part, the security of different NR-VQA models on adversarial videos is evaluated under both white-box and black-box settings, and the results are given in Tables \ref{white-box} and \ref{black-box}, respectively. It can be observed that SecureVQA demonstrates significant advancements in model security while achieving superior VQA performance. For the white-box setting, SecureVQA achieves optimal robustness across all the testing sets. Specifically, SecureVQA improves the robustness by an impressive 110.7\% and 120.0\% on SRCC and PLCC averaged on the above five testing sets compared with HVS-5M \cite{HVS-5M}. Meanwhile, the SRCC of SecureVQA is only degraded by 27.4\% on $\text{LSVQ}_\text{1080P}$ after being attacked. Considering that the parameters and structure of the model are known to the attacker under the white-box setting, it is straightforward for the attacker to attack the VQA model through backpropagation. Therefore, SecureVQA achieves impressive effectiveness in the white-box scenario. For the black-box setting,  SecureVQA provides complete resistance to existing attack methods. Specifically, the SRCC and PLCC of SecureVQA are only slightly disturbed by 0.8\% and 0.8\% on the above five testing sets, which validates the effectiveness of our proposed SecureVQA.

To further validate the security of SecureVQA and its ability to maintain the imperceptibility of the perturbations in the original video, we present some examples of the predicted scores of the original video frames and the corresponding adversarial video frames in Fig. \ref{Effect_picture}. It can be observed that SecureVQA offers formidable security against both white-box and black-box attacks. Specifically, SecureVQA not only prevents high-quality videos from being rated as low quality, it also works the other way around. Furthermore, the modifications introduced by SecureVQA are virtually undetectable, preserving the detailed information and the quality of the video content to the fullest extent.

\begin{table*}[t]
\centering
\caption{Ablation studies on KoNViD-1k. Here, the performance before the attack is marked in gray, and the performance improvement and degradation are colored in \textcolor{red}{red} and \textcolor{blue}{blue}, respectively.}\label{Ablation-Ko}
\renewcommand\arraystretch{1.33}
\fontsize{9}{9}\selectfont
\setlength{\tabcolsep}{0.07mm}{
\begin{tabular}{c:cccc:cccc:cccc:cccc}
\hline
\multirow{2}{*}{Robustness strategy}                  & \multicolumn{4}{c:}{\multirow{2}{*}{None}}                                                                                                                & \multicolumn{12}{c}{SecureVQA (Ours)}                                                                                                                                                                                                                                                                                                                                                                                                                                  \\ \cline{6-17}
                                                      & \multicolumn{4}{c:}{}                                                                                                                                     & \multicolumn{4}{c}{Pixel-wise randomization}                                                                                                                            & \multicolumn{4}{c}{Temporal information}                                                                                                                & \multicolumn{4}{c}{Spatial grid sampling}                                                                                                  \\ \hline
Metric                            &  & SRCC    & PLCC    & $R$       &  & SRCC       & PLCC       & $R$      &  & SRCC           & PLCC           & $R$          &  & SRCC        & PLCC        & $R$       \\ \hline
\multirow{2}{*}{VSFA \cite{VSFA}}             &  & -0.0305 & 0.0586  & 1.6573 &  &     0.7134       &           0.7660 &    4.5038    &  &          0.1858      &     0.2687           &           1.9029 &  &             0.3801    &      0.3408          &           3.2400        \\
                                  &  & \textcolor{gray}{0.7882}  & \textcolor{gray}{0.8106}  & \textcolor{gray}{-}       &  &            \textcolor{gray}{0.7804}  & \textcolor{gray}{0.7980}  & \textcolor{gray}{-}        &  &                \textcolor{gray}{0.8387}  & \textcolor{gray}{0.8250}  & \textcolor{gray}{-}            &  &              \textcolor{gray}{0.6043}  & \textcolor{gray}{0.5364}  & \textcolor{gray}{-}        \\ \hline
\multirow{2}{*}{MDTVSFA \cite{MDTVSFA}}          &  &         0.0261  & 0.1235  & 1.7587         &  &     0.7808       &        0.7816    &   4.9392     &  &       0.6739         &         0.6651       &     2.5890       &  &      0.4670       &    0.4550         & 3.6491        \\
                                  &  &         \textcolor{gray}{0.8003}  & \textcolor{gray}{0.8074}  & \textcolor{gray}{-}         &  &            \textcolor{gray}{0.8162}  & \textcolor{gray}{0.8111}  & \textcolor{gray}{-}       &  &                \textcolor{gray}{0.8335}  & \textcolor{gray}{0.8175}  & \textcolor{gray}{-}           &  &             \textcolor{gray}{0.6180}  & \textcolor{gray}{0.5918}  & \textcolor{gray}{-}          \\ \hline
\multirow{2}{*}{TiVQA \cite{TiVQA}}            &  &        -0.4992 & -0.4099 & 1.2458         &  &           0.7963 &   0.8300         & 4.6232       &  &           -0.1892     &    -0.1291            &  1.5743          &  &        0.4570     &     0.4141        &    3.4416     \\
                                  &  &         \textcolor{gray}{0.8046}  & \textcolor{gray}{0.8377}  & \textcolor{gray}{-}         &  &            \textcolor{gray}{0.8013}  & \textcolor{gray}{0.8276}  & \textcolor{gray}{-} &  &                \textcolor{gray}{0.8235}  & \textcolor{gray}{0.8372}  & \textcolor{gray}{-}           &  &             \textcolor{gray}{0.6593}  & \textcolor{gray}{0.6050}  & \textcolor{gray}{-}                 \\ \hline
\multirow{2}{*}{BVQA-2022 \cite{BVQA-2022}}        &  &      -0.1779     & -0.1465  & 1.4982        &  &         0.8208   &     0.8328       &    5.3932    &  &                0.1978  & 0.2215  & 2.0480            &  &      0.5985       &      0.5836 &        5.9795 \\
                                  &  &        \textcolor{gray}{0.8288}  & \textcolor{gray}{0.8307}  & \textcolor{gray}{-}        &  &            \textcolor{gray}{0.8147}  & \textcolor{gray}{0.8319}  & \textcolor{gray}{-}       &  &                \textcolor{gray}{0.8427}  & \textcolor{gray}{0.8469}  & \textcolor{gray}{-}           &  &             \textcolor{gray}{0.6859}  & \textcolor{gray}{0.6400}  & \textcolor{gray}{-}         \\ \hline
\multirow{2}{*}{HVS-5M \cite{HVS-5M}}        &  &      -0.3610     & -0.2723  & 1.3036        &  &         0.8249   &     0.8395       &    5.8105    &  &                -0.1207 & -0.0404 & 1.5116            &  &      0.4532       &      0.4684 &        3.7098 \\
                                  &  &        \textcolor{gray}{0.8455}  & \textcolor{gray}{0.8562}  & \textcolor{gray}{-}        &  &            \textcolor{gray}{0.8335}  & \textcolor{gray}{0.8434}  & \textcolor{gray}{-}       &  &                \textcolor{gray}{0.8715}  & \textcolor{gray}{0.8605}  & \textcolor{gray}{-}                  &  &             \textcolor{gray}{0.5627}  & \textcolor{gray}{0.5554}  & \textcolor{gray}{-}         \\ \hline
Improvement in robustness &  &      -   &     -    &    -     &  &   \textcolor{red}{+122.6\%}         &     \textcolor{red}{+113.6\%}               &   \textcolor{red}{+245.9\%}     &  &   \textcolor{red}{+43.3\%}             &               \textcolor{red}{+39.1\%} &      \textcolor{red}{+28.2\%}      &  &     \textcolor{red}{+100.5\%}        &       \textcolor{red}{+92.0\%}      &     \textcolor{red}{+172.6\%}    \\ \hline
Change in performance     &  &      -   &     -    &    -     &  &   \textcolor{blue}{-0.5\%}         &    \textcolor{blue}{-0.8\%}                &  -      &  &    \textcolor{red}{+3.5\%}            &      \textcolor{red}{+1.1\%}          &           - &  &     \textcolor{blue}{-23.0\%}        &      \textcolor{blue}{-29.3\%}        &  -       \\ \hline

\end{tabular}}
\end{table*}
\begin{table*}[t]
\centering
\caption{Ablation studies on LIVE-VQC. Here, the performance before the attack is marked in gray, and the performance improvement and degradation are colored in \textcolor{red}{red} and \textcolor{blue}{blue}, respectively. } \label{Ablation-VQC}
\renewcommand\arraystretch{1.33}
\fontsize{9}{9}\selectfont
\setlength{\tabcolsep}{0.2mm}{
\begin{tabular}{c:cccc:cccc:cccc:cccc}
\hline
\multirow{2}{*}{Robustness strategy}                  & \multicolumn{4}{c:}{\multirow{2}{*}{None}}                                                                                                                & \multicolumn{12}{c}{SecureVQA (Ours)}                                                                                                                                                                                                                                                                                                                                                                                                                                  \\  \cline{6-17}
                                                      & \multicolumn{4}{c:}{}                                                                                                                                     & \multicolumn{4}{c}{Pixel-wise randomization}                                                                                                                            & \multicolumn{4}{c}{Temporal information}                                                                                                                & \multicolumn{4}{c}{Spatial grid sampling}                                                                                                  \\ \hline
Metric                            &  & SRCC    & PLCC    & $R$       &  & SRCC       & PLCC       & $R$      &  & SRCC           & PLCC           & $R$          &  & SRCC        & PLCC        & $R$       \\ \hline
\multirow{2}{*}{VSFA \cite{VSFA}}             &  & -0.1605 & -0.0132 & 1.3233 &  &     0.6847       &           0.7461 &    3.1917    &  &            0.1966    &      0.4537          &           1.8394 &  &            0.4089    &      0.3725          &           3.2660         \\
                                  &  & \textcolor{gray}{0.7665}  & \textcolor{gray}{0.7506}  & \textcolor{gray}{-}       &  &            \textcolor{gray}{0.7694}  & \textcolor{gray}{0.7946}  & \textcolor{gray}{-}        &  &                \textcolor{gray}{0.7769}  & \textcolor{gray}{0.8055}  & \textcolor{gray}{-}            &  &             \textcolor{gray}{0.5339}  & \textcolor{gray}{0.5440}  & \textcolor{gray}{-}                    \\ \hline
\multirow{2}{*}{MDTVSFA \cite{MDTVSFA}}          &  &         -0.0074 & 0.0807  & 1.4391         &  &     0.7532       &        0.7767    &   4.2710     &  &          0.7338      &      0.7700         &   3.2877       &  &       0.6230      &            0.6201  &  4.2558         \\
                                  &  &         \textcolor{gray}{0.7908}  & \textcolor{gray}{0.8091}  & \textcolor{gray}{-}         &  &            \textcolor{gray}{0.7834}  & \textcolor{gray}{0.7998}  & \textcolor{gray}{-}       &  &                \textcolor{gray}{0.8052}  & \textcolor{gray}{0.8301}  & \textcolor{gray}{-}            &  &             \textcolor{gray}{0.6383}  & \textcolor{gray}{0.6394}  & \textcolor{gray}{-} \\ \hline
\multirow{2}{*}{TiVQA \cite{TiVQA}}            &  &        -0.0762 & 0.0452  & 1.3234         &  &           0.7832 &      0.7618      &    2.9537    &  &          0.1227      &    0.2054            &           1.6931 &  &     0.3628        & 0.3616             &    2.4186     \\
                                  &  &         \textcolor{gray}{0.8179}  & \textcolor{gray}{0.8067}  & \textcolor{gray}{-}         &  &            \textcolor{gray}{0.8201}  & \textcolor{gray}{0.7982}  & \textcolor{gray}{-} &  &                \textcolor{gray}{0.8526}  & \textcolor{gray}{0.7954}  & \textcolor{gray}{-}            &  &             \textcolor{gray}{0.6570}  & \textcolor{gray}{0.6310}  & \textcolor{gray}{-}         \\ \hline
\multirow{2}{*}{BVQA-2022 \cite{BVQA-2022}}        &  &         0.4034  & 0.4215  & 2.0021        &  &         0.7032   &     0.7806       &    5.8575    &  &                0.4571  & 0.4778  & 2.2061            &  &       0.5995      &     0.5814        &   6.2096      \\
                                  &  &        \textcolor{gray}{0.8041}  & \textcolor{gray}{0.8141}  & \textcolor{gray}{-}        &  &            \textcolor{gray}{0.7946}  & \textcolor{gray}{0.8088}  & \textcolor{gray}{-}       &  &                \textcolor{gray}{0.8698}  & \textcolor{gray}{0.8457}  & \textcolor{gray}{-}           &  &        \textcolor{gray}{0.6633}     &            \textcolor{gray}{0.6744} &     \textcolor{gray}{-}    \\ \hline
\multirow{2}{*}{HVS-5M \cite{HVS-5M}}        &  &      -0.0844     & 0.1798  & 1.7006        &  &         0.7771   &     0.8231       &    5.1940    &  &                0.1948  & 0.3045   & 1.6168            &  &      0.5770       &      0.6276 &        3.4684 \\
                                  &  &        \textcolor{gray}{0.8142}  & \textcolor{gray}{0.8277}  & \textcolor{gray}{-}        &  &            \textcolor{gray}{0.7806}  & \textcolor{gray}{0.8280}  & \textcolor{gray}{-}       &  &                \textcolor{gray}{0.8662}  & \textcolor{gray}{0.8802}  & \textcolor{gray}{-}           &  &             \textcolor{gray}{0.6938}  & \textcolor{gray}{0.7248}  & \textcolor{gray}{-}         \\ \hline
Improvement in robustness &  &      -   &     -    &    -     &  &     \textcolor{red}{+92.0\%}       &    \textcolor{red}{+79.0\%}        &    \textcolor{red}{+124.3\%} &  &     \textcolor{red}{+39.5\%}            &          \textcolor{red}{+35.9\%}      &     \textcolor{red}{+40.1\%}            &  &            \textcolor{red}{+78.9\%} &       \textcolor{red}{+61.7\%}      &  \textcolor{red}{+147.9\%}       \\ \hline
Change in performance     &  &      -   &     -    &    -     &  &    \textcolor{blue}{-1.1\%}        &     \textcolor{red}{+0.6\%}              &  -      &  &     \textcolor{red}{+4.4\%}                  &    \textcolor{red}{+3.7\%}            &      -      &  &             \textcolor{blue}{-20.3\%}&      \textcolor{blue}{-20.0\%}       &  -      \\ \hline

\end{tabular}}
\end{table*}

\subsection{Evaluations on quality prediction} \label{categorical subsets}

In this part, the performance of SecureVQA on clean videos is evaluated on the large-scale LSVQ dataset. Besides, it is further fine-tuned on the small-scale datasets, and the results are given in Tables \ref{LSVQ} and \ref{fine-tuning}, respectively. It can be seen that SecureVQA achieves competitive performance compared with SOTA VQA models. Specifically, SecureVQA has consistently secured a position within the top three across all test categories. Meanwhile, its remarkable performance on the LIVE-VQC dataset, which encompasses rich motion information, can be attributed to SecureVQA's meticulous consideration of inter-frame information and its effective fusion of embeddings from two distinct branches. 
Considering that SecureVQA takes security as
a prerequisite and then ensures its exceptional VQA performance.
As a result, certain strategies like the guardian map may have
a detrimental effect on performance due to the inherent tradeoff between security and accuracy (as shown in the ablation study). Despite this, SecureVQA exhibits an accuracy difference of merely 1\% to 2\% compared to the most advanced DOVER \cite{DOVER}.
The results from both this part and the previous part confirm that SecureVQA has successfully achieved the expected goal depicted in Fig. \ref{Definition_SecureVQA}. It not only demonstrates a formidable resistance against adversarial attacks but also attains superior VQA performance.


\begin{table}[]
\centering
\caption{Performance evaluations under black-box setting on FAST-VQA \cite{FASTVQA}.}\label{FASTVQA-black-box}
\renewcommand\arraystretch{1.33}
\fontsize{9}{9}\selectfont
\setlength{\tabcolsep}{0.40mm}{
\begin{tabular}{c:cccc:cccc}
\hline
Dataset                    &           & \multicolumn{3}{c:}{KoNViD-1k}     &  & \multicolumn{3}{c}{LIVE-VQC} \\  \cmidrule(lr){3-5} \cmidrule(lr){7-9} 
Metric                     &           & SRCC      & PLCC      &          &  & SRCC      & PLCC     & $R$      \\ \hline
\multirow{2}{*}{FAST-VQA \cite{FASTVQA}}      &           & 0.8234 & 0.8218 & 5.0867    &  & 0.7983 & 0.7971 & 3.5765     \\
                           &  & \textcolor{gray}{0.8779}  & \textcolor{gray}{0.8679}  & \textcolor{gray}{-} &  & \textcolor{gray}{0.8477}  & \textcolor{gray}{0.8513}  & \textcolor{gray}{-}  \\ \hline
\multirow{2}{*}{SecureVQA (Ours)}      &           & \textbf{0.8788}     &     \textbf{0.8590}       & \textbf{5.7787}           &  &       \textbf{0.8635}     &  \textbf{0.8669}         &  \textbf{4.2794}         \\
                           &           &           \textbf{\textcolor{gray}{0.8805}}          &          \textbf{\textcolor{gray}{0.8692}} &    \textcolor{gray}{-}           &  &           \textbf{\textcolor{gray}{0.8861}}          &          \textbf{\textcolor{gray}{0.8959}} &    \textcolor{gray}{-}  \\ \hline
\end{tabular}}
\end{table}

\begin{table}[t]
\centering
\caption{Performance evaluations compared with concatenation.}\label{fusion}
\renewcommand\arraystretch{1.33}
\fontsize{9}{9}\selectfont
\setlength{\tabcolsep}{1mm}{
\begin{tabular}{c:ccc:ccc}
\hline
Dataset                    &           & \multicolumn{2}{c:}{KoNViD-1k}     &  & \multicolumn{2}{c}{LIVE-VQC} \\  \cmidrule(lr){3-4} \cmidrule(lr){6-7} 
Metric                     &           & SRCC      & PLCC                &  & SRCC      & PLCC       \\ \hline
Concatenation     &           & 0.8810 & 0.8781     &  & 0.8788 & 0.8764      \\
                         
Fusion manner (SecureVQA)     &           & \textbf{0.8853}     &     \textbf{0.8834}                  &  &       \textbf{0.8817}     &  \textbf{0.8803}                  
                              \\ \hline
\end{tabular}}
\end{table}

\subsection{Ablation study} \label{mixed datasets}
In this part, ablation studies are expected to yield valuable insights into the selection of strategies for future model designs. We investigate the individual contributions of three strategies in SecureVQA to security, namely spatial grid sampling, pixel-wise randomization, and temporal information extraction. The experiments in this subsection are performed under the more practical black-box setting.

\begin{figure*}[t]
	\centering
	\includegraphics[width=\textwidth]{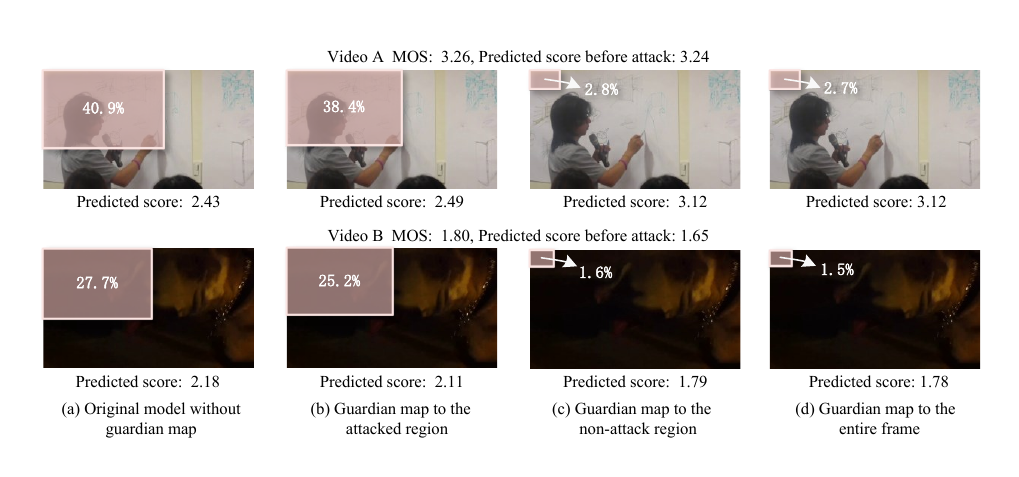}
	\caption{The defense effect of the guardian map added in different regions and corresponding predicted scores. Here, the pink area denotes the size of the disturbed area after the attack. 
}\label{Guardian}
\end{figure*}


Firstly, the generalization of these three strategies to existing models is verified. Each strategy is individually incorporated into the original model in a plug-and-play manner to verify its defense effect and characteristics. Specifically, 1) Inter-frame embeddings can be easily concatenated with the embeddings of existing models. 2) The existing models typically feed pre-processed frames into the networks, thus we simply need to substitute them to the frames processed by spatial grid sampling. 3) Pixel-wise randomization only entails initializing the guardian map of the same size as the input frame and adding it pixel-wise. The experiments are conducted on two widely used medium-sized datasets KoNViD-1k and LIVE-VQC, and the results are given in Tables \ref{Ablation-Ko} and \ref{Ablation-VQC}. From the above two Tables, the following observations can be made.

\begin{table}[t] 
\centering
\caption{The runtime of the different branches and their defenses effect on LIVE-VQC under the black-box setting.}\label{runtime}
\renewcommand\arraystretch{1.4}
\fontsize{9}{9}\selectfont
\setlength{\tabcolsep}{2.35mm}{
\begin{tabular}{c:c:c:c}
\hline
LIVE-VQC                                        & SRCC   & PLCC   & \makecell[c]{Runtime \\per frame (s)}     \\ \hline
\multirow{2}{*}{Intra branch}                    & 0.8215 & 0.8174 & \multirow{2}{*}{\textbf{0.00045}} \\
                                                 & \textcolor{gray}{0.8568} & \textcolor{gray}{0.8601} &                          \\ \hline
\multirow{2}{*}{\makecell[c]{SecureVQA \\(Intra\&Inter branch)} } & \textbf{0.8635} & \textbf{0.8669} & \multirow{2}{*}{0.369}   \\
                                                 & \textcolor{gray}{\textbf{0.8861}} & \textcolor{gray}{\textbf{0.8959}} &                          \\ \hline
\end{tabular}}

\end{table}

1) Pixel-wise randomization exhibits incredible contribution to robustness across all models, yet has minimal impact on model performance. Specifically, compared with the original models, the pixel-wise randomization improves the robustness by a remarkable 107.3\% on SRCC averaged on the above two datasets, while the performance degradation is only 0.8\% on SRCC, demonstrating the effectiveness and versatility of our proposed strategy. 2) Temporal information extraction also plays a key role in model robustness,  and it is the sole strategy that contributes to improving model performance. Specifically, incorporating inter-frame information enhances performance by 4.0\% and boosts robustness by 41.4\% on SRCC averaged on the above two datasets. This substantiates the significance of inter-frame information in the VQA task from an alternative perspective. 3) Spatial grid sampling also demonstrates remarkable defense effects while substantially improving training efficiency. However, a substantial proportion of the video content is discarded during the sampling process, which leads to notable performance degradation when applied to other models. Specifically, applying spatial sampling to other models brings 21.7\% performance degradation on SRCC averaged on the above two datasets.

Besides, it can be seen from Tables \ref{Ablation-Ko} and \ref{Ablation-VQC} that certain strategies may have a detrimental effect on performance due to the inherent tradeoff between security on adversarial videos and accuracy on clean videos. To ensure the security of SecureVQA as well as its accuracy, we adopt three strategies to optimize this trade-off. 1) We employ spatial grid sampling in the intra-branch to enhance the model's robustness. However, this may result in the loss of some global information. Therefore, we leverage the entire frame information in the inter-branch to improve performance on clean videos. 2) We set the value of guardian map to +1 or -1, which can not only effectively mitigate the attack effect on adversarial videos but also have minimal alterations to clean videos. 3) We extract inter-frame information to improve model performance on clean videos, while resisting the impact of intra-frame perturbations on robustness. In summary, the experiments in this part demonstrate the versatility of the proposed framework, showcasing its ease of application to other models. Additionally, it empowers future model designers to select different strategies based on their specific requirements.

Secondly, the mechanism of the pixel-wise randomization through guardian maps to the robustness of the model is further explored. The experiments are conducted via attacking VSFA~\cite{VSFA} on the KoNViD-1k. 
We adopt the only black-box attack method \cite{AttackVQA}, which tentatively perturbs a portion of the frame while leaving the remainder untouched. Therefore, the proportion of the disturbed region (i.e. pink area in Fig. \ref{Guardian}) 
indicates the effectiveness of the pixel-wise randomization for the model's robustness, i.e. smaller disturbed area corresponds to higher robustness.
During the attack process, guardian maps (GMs) are respectively added to the attacked region, the non-attack region, and the entire frame to investigate the areas where guardian maps provide effective defense. From Fig. \ref{Guardian}, three results can be obtained. 1) From (a) and (b), it can be seen that the guardian map neutralizes the carefully designed adversarial perturbations so that the attack is compromised to some extent. 2) From  (a) and (c), it can be observed that the most significant contribution of the guardian map comes from its modifications in the non-attack region. This can be explained by the fact that adding a guardian map to a non-attack region can confuse the attacker whether the perturbations or the guardian map are contributing to the adversarial effect at each query. Besides, it can also provide global immunity to the model, ensuring robustness against local perturbations in advance. As a consequence, the model exposes fewer weaknesses to malicious queries from adversarial attackers. 3) From (a) to (d), the phenomenon described in the above two cases coexists when the guardian map is added to the entire frame. It further diminishes the disturbed area, and thus disables the attack effect, attaining a desirable result close to the score before the attack. 

Thirdly, some methods like FAST-VQA \cite{FASTVQA} also adopt spatial grid sampling. In Table \ref{FASTVQA-black-box}, we verify its defense effect compared with SecureVQA.  It can be observed that SecureVQA exhibits greater robustness compared with \cite{FASTVQA}. Specifically, SecureVQA improves the average robustness by 4.7\% and 3.6\% on SRCC and PLCC compared with \cite{FASTVQA}. This can be attributed to the defensive capabilities of strategies like the guardian map and inter-frame information in SecureVQA. Furthermore, the primary contribution of this paper is not spatial grid sampling, as only adopting the guardian map can provide better defense effects (as shown in Tables \ref{Ablation-Ko} and \ref{Ablation-VQC}). The primary concern of this paper is to design and discover some strategies tailored to the VQA and can be easily integrated into existing models, enabling the model to exhibit defense capabilities without compromising performance on clean videos. Additionally, different strategies in SecureVQA have different characteristics (as shown in Tables \ref{Ablation-Ko} and \ref{Ablation-VQC}), which empower future model designers to select different strategies based on their specific requirements. 

Fourthly, existing VQA models typically employ straightforward concatenation to integrate different types of embeddings. However, we believe that such concatenation does not fully harness the potential of these sophisticated embeddings. Therefore, SecureVQA designs a fusion module to fuse these embeddings extracted by different kinds of backbone networks through some trainable parameters. Table \ref{fusion} displays the performance of adopting the fusion module in SecureVQA on clean videos compared with concatenation. It can be observed that after adopting the fusion manner, the performance on these two datasets has been improved to some extent, demonstrating that the fusion manner in SecureVQA can enhance the representation capability of the different types of embeddings.

\subsection{Computational complexity of the defense strategy}
The strategies in SecureVQA have almost no impact on computational complexity and can even speed up training and inference processes compared with previous works. For instance, many previous studies try to extract intra-frame and inter-frame information through the frame with the original resolution \cite{VSFA,BVQA-2022,HVS-5M}. However, In the intra-frame branch in SecureVQA, we employ spatial grid sampling to reduce the resolution of the input frame. In the inter-frame branch, SecureVQA adopts a resize operation to reduce the resolution of the input frame and then extracts the motion information. Moreover, the guardian map also introduces almost no computational complexity. Therefore, the three defense strategies in SecureVQA do not impose significant computational complexity while maintaining effective defense properties.

Besides, SecureVQA designs two branches to handle intra- and inter-frame information, respectively. However, SecureVQA utilizes a relatively large network SlowFast \cite{Slowfast} to extract inter-frame information, which requires offline processing due to its higher computational cost. In Table \ref{runtime}, we present the defense effect and the runtime of intra-branch and SecureVQA (intra- $\&$ inter-branch) on a single 4090 card under the black-box setting. The experiments are conducted via attacking on the LIVE-VQC dataset, which contains 1080P resolution videos. It can be observed that the runtime of SecureVQA is acceptable when assessing videos with high-resolution videos of 1080P. Moreover, if real-time conditions are the primary focus, only the use of intra-branch can also get a satisfactory defense effect. In the future, we may explore the design of a more lightweight network to handle inter-frame information, enhancing its robustness against adversarial videos while maintaining real-time requirements.

\section{Conclusions}
In this paper, we perform a pioneering exploration into the security of VQA and focus on general adversarial defense principles, aiming to endow existing VQA models with security. Building upon our exploration, we propose a secure VQA framework named SecureVQA, which is seemingly simple but can completely resist the current adversarial attack methods against the NR-VQA models while achieving superior VQA performance. Extensive experiments have been conducted, and the following conclusions can be drawn. Firstly, randomness serves as a formidable opponent to the optimization-based attack algorithms. Therefore, we introduce two principles to increase the security of the model for the intra-frame defense, namely spatial grid sampling and pixel-wise randomization. Secondly, temporal information is a strong guarantee against intra-frame adversarial perturbations. Therefore, we extract temporal embeddings for the inter-frame defense. The ablation study demonstrates that the principles in our framework can serve as a general defense component for leading VQA models. These findings from SecureVQA are expected to yield valuable insights for the establishment of a secure VQA paradigm.

It is worth noting that the vulnerability of QA models has attracted more and more attention in the last two years. This paper explored the general defense principles from the intra- and inter-frame perspectives. Our empirical results show that the intra-frame defense performs better than the inter-frame defense. This is because current adversarial attacks only focus on spatial perturbation, and no attempt has been made to attack the temporal domain of motion video so far. To this end, we extract the temporal information as a complementary safeguard. However, it might be invalid to the next generation of VQA where the motion information could be exploited for an adversarial attack in the future. As we predicted, applying temporal smoothing or filtering, similar to the pixel-wise randomization, might mitigate such attacks. Besides, our two principles (i.e. spatial grid sampling and pixel-wise randomization) contribute to the security of VQA models the most, but, are also at the cost of a slight degradation in VQA accuracy. How to efficiently exploit these two characteristics needs more detailed and deeper exploration. Looking back on widespread social effects, we are the first to defend the adversarial attack on VQA models.  Targeting our defense principles, a new round of adversarial attacks could be triggered. Hence, such a ``cat-mouse" game has emerged in the VQA field, resulting in a never-ending battle. Despite this, we believe that SecureVQA paves a new avenue toward establishing a reliable and practical QA system since these general defense principles outlined in our framework can be seamlessly integrated into existing VQA models.


\begin{thebibliography}{5}
\bibliographystyle{ieeetr} \small 

\bibitem{Trafficreport}
NCTA. Where does the majority of internet traffic come from? [Online] Available: https://www.ncta.com/whats-new/
report-where-does-the-majority-of-internet-traffic-come/, 2019.

\bibitem{TOB1}
T. Li, X. Min, H. Zhao, G. Zhai, Y. Xu, and W. Zhang, ``Subjective and objective quality assessment of compressed screen content videos,'' \emph{IEEE Transactions on Broadcasting}, vol. 67, no. 2, pp. 438-449, 2021.

\bibitem{TOB5}
T. Zhou, S. Tan, W. Zhou, Y. Luo, Y.-G. Wang, and G. Yue, ``Adaptive mixed-scale feature fusion network for blind AI-generated image quality assessment," \emph{IEEE Transactions on Broadcasting}, vol. 70, no. 3, pp. 833-843, 2024.

\bibitem{TOB3}
S. Jiang, Q. Sang, Z. Hu, and L. Liu, ``Self-supervised representation learning for video quality assessment,'' \emph{IEEE Transactions on Broadcasting}, vol. 69, no. 1, pp. 118-129, 2023.

\bibitem{TOB4}
F. Xing, M. Li, Y.-G. Wang, G. Zhu, and X. Cao, ``CLIPVQA: Video quality assessment via CLIP,"  \emph{IEEE Transactions on Broadcasting}, vol. 71, no. 1, pp. 291-306, 2025. 

\bibitem{AttackIQA}
W. Zhang, D. Li, X. Min, G. Zhai, G. Guo, X. Yang, and K. Ma, ``Perceptual attacks of no-reference image quality models with human-in-the-loop,'' in \emph{Advances in Neural Information Processing Systems (NeurIPS)}, pp. 2916-2929, 2022.

\bibitem{BMVC_AttackVQA}
E. Shumitskaya, A. Antsiferova, and D. Vatolin, ``Universal perturbation attack on differentiable no-reference image- and video-quality metrics,'' in \emph{British Machine Vision Conference (BMVC)}, pp. 1-12, 2022.

\bibitem{AttackVQA}
A.-X. Zhang, Y. Ran, W. Tang, and Y.-G. Wang, ``Vulnerabilities in video quality assessment models: The challenge of adversarial attacks,'' in \emph{Advances in Neural Information Processing Systems (NeurIPS)}, pp. 1-12, 2023.

\bibitem{Defense_1}
J. Korhonen and J. You, ``Adversarial attacks against blind image quality assessment models,'' in \emph{second Workshop on Quality of Experience in Visual Multimedia Applications (QoEVMA)}, pp. 1-9, 2022.

\bibitem{Defense_2}
A. Gushchin, A. Chistyakova, V. Minashkin, A. Antsiferova, and D. Vatolin, ``Adversarial purification for no-reference image-quality metrics: Applicability study and new methods,'' \emph{arXivpreprint arXiv}: 2404.06957, 2024.

\bibitem{AttackIQA_CSVT}
C. Yang, Y. Liu, D. Li, and T. Jiang, ``Exploring vulnerabilities of no-reference image quality assessment models: A query-based black-box method,'' \emph{IEEE Transactions on Circuits and Systems for Video Technology}, vol. 34, no. 12, pp. 12715-12729, 2024.

\bibitem{Ti-Patch}
V. Leonenkova, E. Shumitskaya, A. Antsiferova, and D. Vatolin, ``Ti-patch: Tiled physical adversarial patch for no-reference video quality metrics,'' \emph{arXivpreprint arXiv}:2404.09961, 2024.

\bibitem{Ranyu}
Y. Ran, A.-X. Zhang, M. Li, W. Tang, and Y.-G. Wang, ``Black-box adversarial attacks against image quality assessment models,'' \emph{Expert Systems with Applications}, vol. 260 (125415), 2025.

\bibitem{Gushchin2024}
A. Gushchin, A. Chistyakova, V. Minashkin, A. Antsiferova, and D. Vatolin, ``Adversarial purification for no-reference image-quality metrics: applicability study and new methods,'' \emph{arXivpreprint arXiv}:2404.06957, 2024.

\bibitem{Q-Bench}
H. Wu, Z. Zhang, E. Zhang, C. Chen, L. Liao, A. Wang, C. Li, W. Sun, Q. Yan, G. Zhai, and W. Lin, ``Q-bench: A benchmark for general-purpose foundation models on low-level vision,'' \emph{arXivpreprint arXiv}:2309.14181, 2023.

\bibitem{Q-Align}
H. Wu, Z. Zhang, W. Zhang, C. Chen, L. Liao, C. Li, Y. Gao, A. Wang, E. Zhang, W. Sun, Q. Yan, X. Min, G. Zhai, and W. Lin, ``Q-align: Teaching lmms for visual scoring via discrete text-defined levels,'' \emph{arXivpreprint arXiv}:2312.17090, 2023.

\bibitem{Yang2024}
C. Yang, Y. Liu, D. Li, Y. Zhong, and T. Jiang, ``Beyond score changes: Adversarial attack on no-reference image quality assessment from two perspectives,'' \emph{arXivpreprint arXiv}:2404.13277, 2024.

\bibitem{Shumitskaya2024}
E. Shumitskaya, A. Antsiferova, and D. Vatolin, ``Towards adversarial robustness verification of no-reference image- and video-quality metrics,'' \emph{Computer Vision and Image Understanding}, vol. 240 (103913), 2024.

\bibitem{Shukla}
A. Shukla, A. Upadhyay, S. Bhugra, and M. Sharma, ``Opinion unaware image quality assessment via adversarial convolutional variational autoencoder,'' in \emph{IEEE Winter Conference on Applications of Computer Vision (WACV)}, pp. 2153-2163, 2024.

\bibitem{Sang}
Q. Sang, H. Zhang, L. Liu, X. Wu, and A. C. Bovik, ``On the generation of adversarial examples for image quality assessment,'' \emph{The Visual Computer}, vol. 40, pp. 3183–3198, 2024.


\bibitem{VSFA}
D. Li, T. Jiang, and M. Jiang, ``Quality assessment of in-the-wild videos,'' in \emph{ACM International Conference on Multimedia (MM),} pp. 2351-2359, 2019.

\bibitem{FASTVQA}
H. Wu, C. Chen, J. Hou, L. Liao, A. Wang, W. Sun, Q. Yan, and W. Lin, ``FAST-VQA: Efficient end-to-end video quality assessment with fragment sampling,'' in \emph{European Conference on Computer Vision (ECCV)}, pp. 538–554, 2022.

\bibitem{TOB2}
W. Shen, M. Zhou, X. Liao, W. Jia, T. Xiang, and B. Fang, ``An end-to-end no-reference video quality assessment method with hierarchical spatiotemporal feature representation,'' \emph{IEEE Transactions on Broadcasting}, vol. 68, no. 3, pp. 651-660, 2022.
 

\bibitem{HVS-5M}
A.-X. Zhang, Y.-G. Wang, W. Tang, L. Li, and S. Kwong, ``A spatial-temporal video quality assessment method via comprehensive HVS simulation,'' \emph{IEEE Transactions on Cybernetics}, vol. 54, no. 8, pp. 4749-4762, 2024.

\bibitem{StableVQA}
T. Kou, X. Liu, W. Sun, J. Jia, X. Min, G. Zhai, and N. Liu, ``StableVQA: A deep no-reference quality assessment model for video stability,'' in \emph{ACM International Conference on Multimedia (MM),} pp. 1066-1076, 2023.

\bibitem{DOVER}
H. Wu, E. Zhang, L. Liao, C. Chen, J. Hou, A. Wang, W. Sun, Q. Yan, and W. Lin, ``Exploring video quality assessment on user generated contents from aesthetic and technical perspectives,'' in \emph{International Conference on Computer Vision (ICCV),} pp. 20144-20154, 2023.

\bibitem{DiscoVQA}
H, Wu, C, Chen, L, Liao, J, Hou, W, Sun, Q. Yan, and W. Lin, ``DisCoVQA: Temporal distortion-content transformers for video quality assessment,'' \emph{IEEE Transactions on Circuits and Systems for Video Technology}, vol. 33, no. 9, pp. 4840-4854, 2023.

\bibitem{TiVQA}
A.-X. Zhang and Y.-G. Wang, ``Texture information boosts video quality assessment,'' in \emph{IEEE International Conference on Acoustics, Speech and Signal Processing (ICASSP),} pp. 2050-2054, 2022.

\bibitem{MaxVQA}
H. Wu, E. Zhang, L. Liao, C. Chen, J. Hou, A. Wang, W. Sun, Q. Yan, and W. Lin, ``Towards explainable in-the-wild video quality assessment: A database and a language-prompted approach,'' in \emph{ACM International Conference on Multimedia (MM),} pp. 1-12, 2023.

\bibitem{Swin}
Z. Liu, J. Ning, Y. Cao, Y. Wei, Z. Zhang, S. Lin, and H. Hu, ``Video swin transformer,'' in \emph{IEEE Conference on Computer Vision and Pattern Recognition (CVPR),}  pp. 3202-3211, 2022.

\bibitem{KoNViD}
V. Hosu, F. Hahn, M. Jenadeleh, H. Lin, H. Men, T. Szir\'anyi, S. Li, and D. Saupe, ``The konstanz natural video database (KoNViD-1k),'' in \emph{Eleventh International Conference on Quality of Multimedia Experience (QoMEX)}, pp. 1-6, 2017.

\bibitem{VQC}
Z. Sinno and A. C. Bovik, ``Large-scale study of perceptual video quality,'' \emph{IEEE Transactions on Image Processing}, vol. 28, no. 2, pp. 612–627, 2019.

\bibitem{YouTube}
Y. Wang, S. Inguva, and B. Adsumilli, ``YouTube UGC dataset for video compression research,'' in \emph{IEEE International Workshop on Multimedia Signal Processing (MMSP),} pp. 1–5, 2019.

\bibitem{LSVQ}
Z. Ying, M. Mandal, D. Ghadiyaram, and A. C. Bovik, ``Patch-VQ: ‘Patching Up’ the video quality problem,'' in \emph{IEEE Conference on Computer Vision and Pattern Recognition (CVPR),}  pp. 14019–14029, 2021.

\bibitem{Slowfast}
C. Feichtenhofer, H. Fan, J. Malik, and K. He, ``SlowFast networks for video recognition,'' in \emph{International Conference on Computer Vision (ICCV),} pp. 6201–6210, 2019.

\bibitem{Kinetics}
W. Kay, J. Carreira, K. Simonyan, B. Zhang, C. Hillier, S. Vijayanarasimhan, F. Viola, T. Green, T. Back, P. Natsev, M. Suleyman, and A. Zisserman, ``The kinetics human action video dataset,'' \emph{arXivpreprint arXiv}:1705.06950, 2017.


\bibitem{SRCC}
M. Blondel, O. Teboul, Q. Berthet, and J. Djolonga, ``Fast differentiable
sorting and ranking,'' in \emph{International Conference on Machine Learning},
pp. 950–959, 2020.

\bibitem{Yu2024}
Z. Yu, F. Guan, Y. Lu, X. Li, and Z. Chen, ``Video quality assessment based on swin TransformerV2 and coarse to fine strategy,'' \emph{arXivpreprint arXiv}:2401.08522, 2024.

\bibitem{Mitra2024}
S. Mitra and R. Soundararajan, ``Knowledge guided semi-supervised learning for quality assessment of user generated videos,'' in \emph{Association for the Advancement of Artificial Intelligence (AAAI)}, pp. 4251-4260, 2024.

\bibitem{TLVQM}
J. Korhonen, “Two-level approach for no-reference consumer video quality assessment,” \emph{IEEE Transactions on Image Processing}, vol. 28, no. 12, pp. 5923–5938, 2019.

\bibitem{VIDEVAL}
Z. Tu, Y. Wang, N. Birkbeck, B. Adsumilli, and A. C. Bovik, ``UGC-VQA: Benchmarking blind video quality assessment for user generated content,'' \emph{IEEE Transactions on Image Processing}, vol. 30, pp. 4449–4464, 2021.

\bibitem{BVQA-2022}
B. Li, W. Zhang, M. Tian, G. Zhai, and X. Wang, ``Blindly assess quality of in-the-wild videos via quality-aware pre-training and motion perception,'' \textit{IEEE Transactions on Circuits and Systems for Video Technology}, vol. 32, no. 9, pp. 5944-5958, 2022.

\bibitem{ImageNet}
J. Deng, W. Dong, R. Socher, L.-J. Li, K. Li, and  F.-F. Li, ``ImageNet: A large-scale hierarchical image database,'' in \emph{IEEE Conference on Computer Vision and Pattern Recognition (CVPR)}, pp. 248-255, 2009.

\bibitem{MDTVSFA}
D. Li, T. Jiang, and M. Jiang, ``Unified quality assessment of in-the-wild videos with mixed datasets training,'' \emph{International Journal of Computer Vision,} vol. 129, no. 4, pp. 1238-1257, 2021.

\bibitem{ZoomVQA}
K. Zhao, K. Yuan, M. Sun, and X. Wen, ``Zoom-VQA: Patches, frames and clips integration for video quality assessment,'' in \emph{IEEE Conference on Computer Vision and Pattern Recognition Workshops (CVPRW)}, pp. 1302-1310, 2023.

\bibitem{MDVQA}
Z. Zhang, W. Wu, W. Sun, D. Tu, W. Lu, X. Min, Y. Chen, and G. Zhai, ``MD-VQA: Multi-dimensional quality assessment for UGC live videos,'' in \emph{IEEE Conference on Computer Vision and Pattern Recognition (CVPR)}, pp. 1746-1755, 2023.

\bibitem{Defense1}
W. Wang, C. Zhou, D. Lin, and Y.-G. Wang, ``FeConDefense: Reversing adversarial attacks via feature consistency loss,'' \emph{Computer Communications}, vol. 211, pp. 263-270, 2023.

\bibitem{Defense3}
C. Mao, M. Chiquier, H. Wang, J. Yang, and C. Vondrick, ``Adversarial attacks are reversible with natural supervision,'' in \emph{International Conference on Computer Vision (ICCV),} pp. 661–671, 2021.

\bibitem{TPQI}
L. Liao, K. Xu, H. Wu, C. Chen, W. Sun, Q. Yan, and W. Lin, ``Exploring the effectiveness of video perceptual representation in blind video quality assessment,'' in \emph{ACM International Conference on Multimedia (MM),} pp. 837-846, 2022.

\bibitem{CoSTA}
F. Xing, Y.-G. Wang, W. Tang, G. Zhu, and S. Kwong, ``CoSTA: Co-training spatial–temporal attention for blind video quality assessment,'' \emph{Expert Systems with Applications,} vol. 255 (124651), 2024.

\bibitem{AdaDqa}
H. Liu, M. Wu, K. Yuan, M. Sun, Y. Tang, C. Zheng, X. Wen, and X. Li, ``Ada-DQA: Adaptive diverse quality-aware feature acquisition for video quality assessment,'' in \emph{ACM International Conference on Multimedia (MM),} pp. 6695-6704, 2023.

\bibitem{SimpleVQA}
W. Sun, X. Min, W. Lu, and G. Zhai, ``A deep learning based no-reference quality assessment model for UGC videos,'' in \emph{ACM International Conference on Multimedia (MM),} pp. 856-865, 2022.

\bibitem{PGD}
A. Madry, A. Makelov, L. Schmidt, D. Tsipras, and A. Vladu, ``Towards deep learning models resistant to adversarial attacks,'' In \emph{International Conference on Learning Representations}, pp. 1-23, 2018. \\ \\


\textbf{Ao-Xiang Zhang} is currently pursuing the M.E. degree at the School of Computer Science and Cyber Engineering, Guangzhou University, Guangzhou, China. His research interests include video quality assessment and AI security.\\

\textbf{Yuan-Gen Wang} received the Ph.D. degree in communication and information systems from Sun Yat-sen University of China in 2013. He is currently a Full Professor at Guangzhou University. His research interests include computer vision and AI security.\\

\textbf{Yu Ran} received the M.S. degree in cyberspace security from Guangzhou University, Guangzhou, China. She is currently pursuing the Ph.D. degree at the National University of Defense Technology, Changsha, China. Her current research interests include AI security, deep learning, and computer vision.\\

\textbf{Weixuan Tang} received the Ph.D. degree in communication and information systems from Sun Yat-sen University of China in 2019. He is currently an Associate Professor at Guangzhou University, Guangzhou. His research interests include digital image steganography, digital image forensics, adversarial examples, and model security.\\

\textbf{Qingxiao Guan} received the Ph.D. degree in control theory from the University of Science and Technology of China, Hefei, China. He is currently an Associate Professor at Guangzhou University, Guangzhou. His research interests include information hiding and multimedia security.\\

\textbf{Chunsheng Yang} received the Ph.D. degree from Hiroshima University, Hiroshima, Japan, in 1995. He is working at Guangzhou University of China as a Full Professor. His research interests include machine learning, prognostic health management, and reasoning technologies. He is a Fellow of the Canadian Academy of Engineering.

\end{thebibliography}
\end{document}